\documentclass[pdflatex,bst-files/sn-mathphys-num]{sn-jnl}

\usepackage{drago}
\usepackage{fix-cm}
\usepackage{hyperref}

\begin{document}

\title[Article Title]{An Algorithmic Approach for Causal Health Equity: A Look at Race Differentials in Intensive Care Unit (ICU) Outcomes}

\author*[1]{\fnm{Drago} \sur{Ple\v cko}}\email{dp3144@columbia.edu}
\author[2]{\fnm{Paul} \sur{Secombe}}\email{paulsecombe@bigpond.com}
\author[3]{\fnm{Andrea} \sur{Clarke}}\email{andrea.clarke@unimelb.edu.au}
\author[4]{\fnm{Amelia} \sur{Fiske}}\email{a.fiske@tum.de}
\author[5]{\fnm{Samarra} \sur{Toby}}\email{nativeacademyofspace@proton.me}
\author[6]{\fnm{Donisha} \sur{Duff}}\email{ceo@qibn.com.au}
% senior authors
\author[7]{\fnm{David} \sur{Pilcher}}\email{d.pilcher@alfred.org.au}
\author[8]{\fnm{Leo Anthony} \sur{Celi}}\email{lceli@mit.edu}
\author[7]{\fnm{Rinaldo} \sur{Bellomo}}\email{rinaldo.bellomo@austin.org.au}
\author[1]{\fnm{Elias} \sur{Bareinboim}}\email{eb@cs.columbia.edu}

\affil[1]{\orgdiv{Department of Computer Science}, \orgname{Columbia University}%, \orgaddress{\street{500W 120th Street}, \city{New York}, \postcode{10027}, \state{NY}, \country{United States}}
}

\affil[2]{\orgdiv{School of Medicine}, \orgname{Flinders University}%, \orgaddress{\street{?}, \city{Bedford Park}, \postcode{?}, \state{SA}, \country{Australia}}
}

\affil[3]{\orgdiv{School of Population and Global Health}, \orgname{University of Melbourne}%, \orgaddress{\street{55 Commercial Rd}, \city{University of Melbourne}, \postcode{3010}, \state{VIC}, \country{Australia}}
}

\affil[4]{\orgdiv{Institute of History and Ethics of Medicine}, 
\orgname{Technical University of Munich}%,\orgaddress{\street{Prinzregentenstraße 68}, \city{Munich}, \postcode{81675}, \country{Germany}}
}

\affil[5]{\orgdiv{Prince Charles Hospital Brisbane}%, \orgname{Edith Cowan University}%, \orgaddress{\street{55 Commercial Rd}, \city{Melbourne}, \postcode{3004}, \state{VIC}, \country{Australia}}
}

\affil[6]{\orgdiv{Kurongkurl Katitjin, Centre for Indigenous Australian Education and Research}, \orgname{Edith Cowan University}%, \orgaddress{\street{55 Commercial Rd}, \city{Melbourne}, \postcode{3004}, \state{VIC}, \country{Australia}}
}

\affil[7]{\orgdiv{Australian and New Zealand Intensive Care Research Centre}, \orgname{Monash University}%, \orgaddress{\street{55 Commercial Rd}, \city{Melbourne}, \postcode{3004}, \state{VIC}, \country{Australia}}
}

\affil[8]{\orgdiv{Laboratory for Computational Physiology}, \orgname{Massachusetts Institute of Technology}%, \orgaddress{\street{77 Massachusetts Avenue}, \city{Boston}, \postcode{02139}, \state{MA}, \country{United States}}
}

% \affil[7]{\orgdiv{Department of Intensive Care}, \orgname{Austin Hospital}%, \orgaddress{\street{145 Studley Rd}, \city{Heidelberg}, \postcode{3084}, \state{VIC}, \country{Australia}}
% }

% ; 
% The , Victoria 3010 Australia
% CARLTON VIC 3010

% \affil[5]{\orgdiv{Division of Pulmonary, Critical Care and Sleep Medicine}, \orgname{Beth Israel Deaconess Medical Center}, \orgaddress{\street{330 Brookline Avenue}, \city{Boston}, \postcode{02215}, \state{MA}, \country{United States}}}

% \affil[6]{\orgdiv{Department of Biostatistics}, \orgname{Harvard T.H. Chan School of Public Health}, \orgaddress{\street{655 Huntington Avenue}, \city{Boston}, \postcode{02215}, \state{MA}, \country{United States}}}

\abstract{
Health equity is defined as the state in which everyone has a fair and just opportunity to attain their highest level of health. Achieving health equity is believed to improve the well-being of communities, reduce healthcare costs, and increase productivity and longevity. However, disparities in health are still significant. In this context, the new era of large-scale data collection and analysis presents an opportunity for diagnosing and understanding the causes of health inequities. In this study, we describe a framework for systematically analyzing health disparities using tools of causal inference. We illustrate the framework by investigating racial and ethnic disparities in intensive care unit (ICU) outcome between majority and minority groups in Australia (Indigenous vs. Non-Indigenous) and the United States (African-American vs. White). We demonstrate that commonly used statistical measures for quantifying inequity are insufficient, and focus on attributing the observed disparity to the causal mechanisms that generate it. We find that minority patients are younger at admission, have worse chronic health, are more likely to be admitted for urgent and non-elective reasons, and have higher illness severity. At the same time, however, we also find a protective direct effect of belonging to a minority group, with minority patients showing improved survival compared to their majority counterparts, with all other variables being equal. We then demonstrate that this protective effect is related to the increased probability of being admitted to ICU, with minority patients having an increased risk of ICU admission. Additionally, we also find that minority patients, while showing improved survival, are in fact more likely to be readmitted to ICU. These findings support the hypothesis that, due to worse access to primary health care, minority patients are more likely to end up in ICU for preventable conditions, causing a reduction in the mortality rates and creating an effect that appears to be protective. Since the baseline risk of ICU admission may serve as proxy for lack of access to primary care, we developed the Indigenous Intensive Care Equity (IICE) Radar, a monitoring system for tracking the over-utilization of ICU resources by the Indigenous population of Australia across geographical areas.
}




\maketitle
\section{Introduction}\label{sec:introduction}
Health equity is defined as the state in which everyone has a fair and just opportunity to attain their highest level of health \citep{braveman2017health}. One of the key goals in pursuing health equity is the elimination of economic, social, and other obstacles to health and health care, together with eliminating existing, preventable differences in health \citep{braveman2014health}. 
%Fostering health equity is crucial for a fair and just society where all individuals, regardless of their background, can lead healthy lives. 
%By focusing on health equity, 
The goal is to reduce preventable illnesses and deaths, improve the overall well-being of communities, and create a more inclusive healthcare system that serves everyone effectively \citep{braveman2011health}. Despite a commitment of public health organizations and government agencies to address issues of health equity \citep{cdc2024healthequity, hhs2024fact, hhs2024advancingequity, who2024health, eu2024health, health2030}, health disparities are still large \citep{commonwealth2023health, DwyerLindgren2022, TheLancetPublicHealth2023}, and many consider the solutions to health inequities to be in their early stages. 
%In other words, society is only starting to solve this issue in a serious manner.

In this context, the rise of the new generation of computational tools and the widespread adoption of electronic health records (EHRs) \citep{evans2016} offer a major opportunity to quantify and understand health disparities. This task is even more important due to a broad transition to using artificial intelligence (AI) tools in healthcare, which may perpetuate or amplify existing biases. In fact, recent works demonstrate formally that understanding disparities in human decision-making is an essential step for understanding disparities in automated AI systems \citep{plevcko2024causal}. 
% The framework that we use in this paper
Therefore, a successful transition to healthcare systems that benefit from the numerous advantages of AI requires careful scrutiny of the data that is given to such AI systems for training, which includes understanding and quantifying the biases in the data itself. In this paper, we demonstrate that significant disparities in health exist in the data even before applying AI tools. Thus, training AI systems using such biased data may lead to further biases, if not handled appropriately \citep{plecko2024mind}. At the same time, while great care needs to be taken when training AI systems, a transition to such systems also presents the opportunity to possibly correct for existing human biases.

So far, a systematic framework for analyzing health equity that is compatible with the new needs of the data-driven era has not been proposed. In this paper, we discuss such a framework that may appeal to a broad range of practitioners, data scientists, and AI engineers. The main aim of the framework is to use a causal lens to obtain actionable insights that can improve patient engagement and outcomes. To illustrate this, we perform a case study and focus on the question of racial and ethnic disparities in outcome following admission to an intensive care unit (ICU) \citep{mcgowan2022racial}, in particular focusing on disparities in mortality. We analyze data from Australia \citep{secombe2023thirty} for disparities between the Indigenous (Aboriginal and Torres Strait Islander) and Non-Indigenous (majority group) population of Australia. In parallel, we also analyze a database from a large tertiary hospital in Boston, Massachusetts \citep{johnson2020mimic}, focusing on the racial disparity between the African-American and White population. 
We illustrate how a systematic approach for analyzing equity can illuminate different causal mechanisms that generate disparities observed in the data, and demonstrate that racial/ethnic disparities in ICU outcome are markedly similar between the United States and Australia. 
%Our results indicate a parallel tale of two countries, in which health disparities are propagated similarly, most likely due to differences in socio-economic well-being and access to healthcare. 

\section{Results}\label{sec:results}
The research in this paper was approved by the ethics committee of the Alfred Hospital, Melbourne (Project \#661/24), and was done in collaboration with the Indigenous Data Network (IDN) of Australia \citep{IDN2024Australia}.

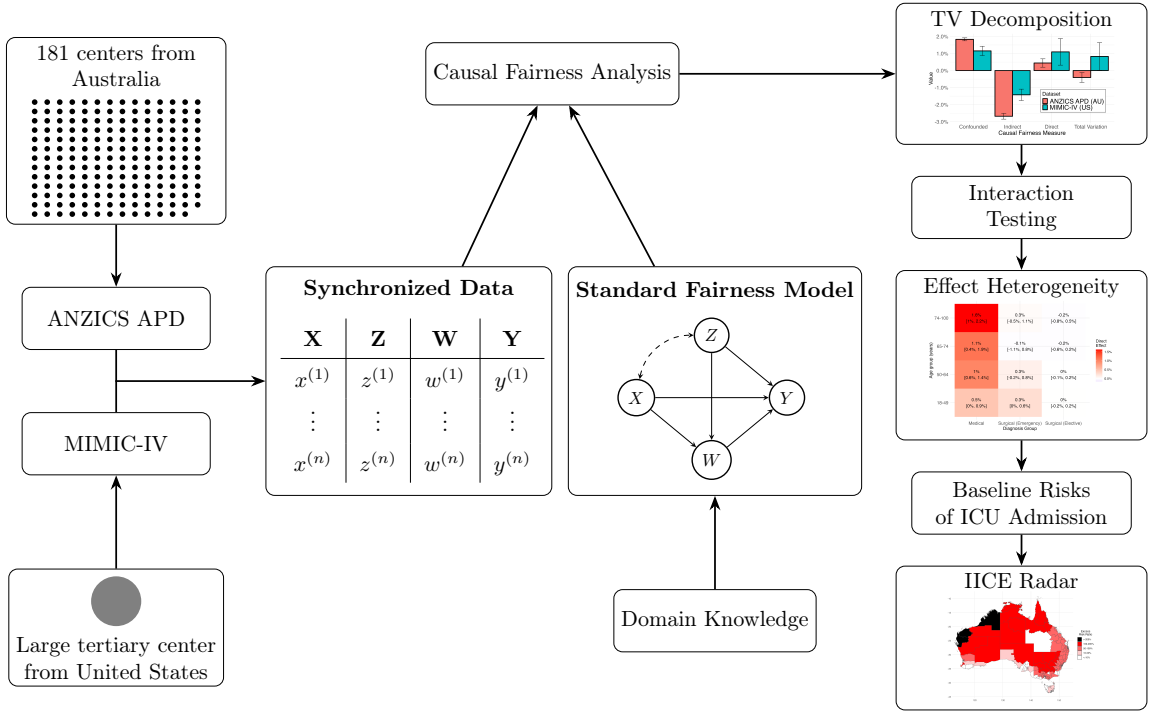
\begin{figure}
    \centering
    \scalebox{0.825}{
        \begin{tikzpicture}[
    node distance=1cm and 1cm,
    box/.style={rectangle, draw, rounded corners, align=center, minimum height=1cm, minimum width=2cm, fill=white},
    arrow/.style={-Stealth, thick},
    rv/.style={circle, draw, thick, minimum size=1cm, align=center},
    ]

    % Leftmost boxes
    \node[box, minimum width = 3.5cm, yshift=0.5cm] (centers) {
    
    181 centers from\\ Australia \\[5pt] 
    \begin{tikzpicture}[scale=0.5]
\foreach \x in {1,...,14} {
  \foreach \y in {1,...,13} {
    \ifthenelse{\x=14 \AND \y=1}{}{%
      \fill[black] ({\x*0.4},{\y*0.3}) circle(0.09cm);
    }
  }
}
\end{tikzpicture}
    };
    \node[box, below=of centers, minimum width = 3cm] (anzics) {ANZICS APD};
    \node[box, below=of anzics, minimum width = 3cm] (mimic) {MIMIC-IV};
    \node[box, below=of mimic,yshift=-0.5cm] (us_center) {
    \begin{tikzpicture}
        \fill[gray] (0, 0) circle(0.4cm); % Large gray dot
    \end{tikzpicture} \\
    Large tertiary center\\ from United States};
    \draw[thick] (anzics) -- (mimic) node[midway] (datmid) {};

    % Data Table - fixed alignment
\node[box, right=of datmid, xshift=1.25cm, minimum width=4.5cm, minimum height=3.675cm, align=center] (data) {%
    \textbf{Synchronized Data} \\[8pt]
    \renewcommand{\arraystretch}{1.5}
    \begin{tabular}{c|c|c|c}
        $\mathbf{X}$ & $\mathbf{Z}$ & $\mathbf{W}$ & $\mathbf{Y}$ \\ \hline
        $x^{(1)}$ & $z^{(1)}$ & $w^{(1)}$ & $y^{(1)}$ \\
        $\vdots$ & $\vdots$ & $\vdots$ & $\vdots$ \\
        $x^{(n)}$ & $z^{(n)}$ & $w^{(n)}$ & $y^{(n)}$ \\
    \end{tabular}
};

    \node[box, right=of data, xshift=-0.75cm, minimum width=4.5cm, minimum height=3.675cm, align=center] (model) {%
        \textbf{Standard Fairness Model} \\[1pt]
        \includegraphics[width=3.5cm]{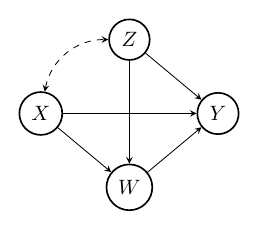}
        \vspace{-0.1in}
    };
    
    % Domain Knowledge
    \node[box, below=of model, yshift=-0.5cm] (domain) {Domain Knowledge};
    
    % Causal fairness analysis
    \node[box, above right=of datmid, xshift=3.8cm, yshift=3.3cm] (causal) {Causal Fairness Analysis};

    % Right side - TV Decomposition and others
    \node[box, right=of causal, xshift=2.45cm, minimum width = 4cm] (tv) {
      TV Decomposition \\
      \includegraphics[width=3cm]{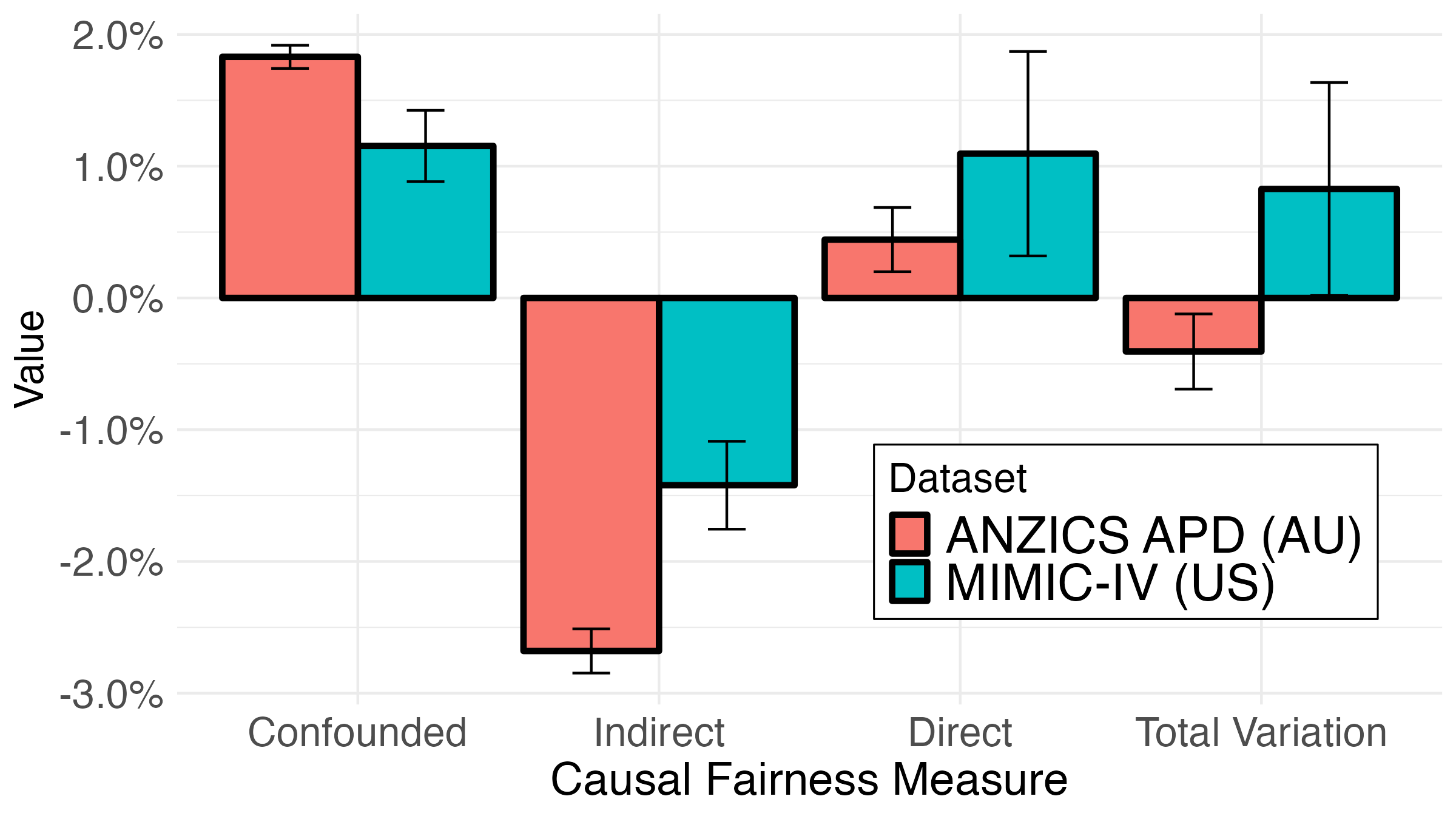}
    };
    \node[box, below=of tv, align=center, yshift=0.5cm, minimum width = 3.5cm] (interaction) {Interaction\\ Testing};
    \node[box, below=of interaction, yshift=0.5cm, align=center, minimum width = 4cm] (heterogeneity) {
      Effect Heterogeneity \\
      \includegraphics[width=3cm]{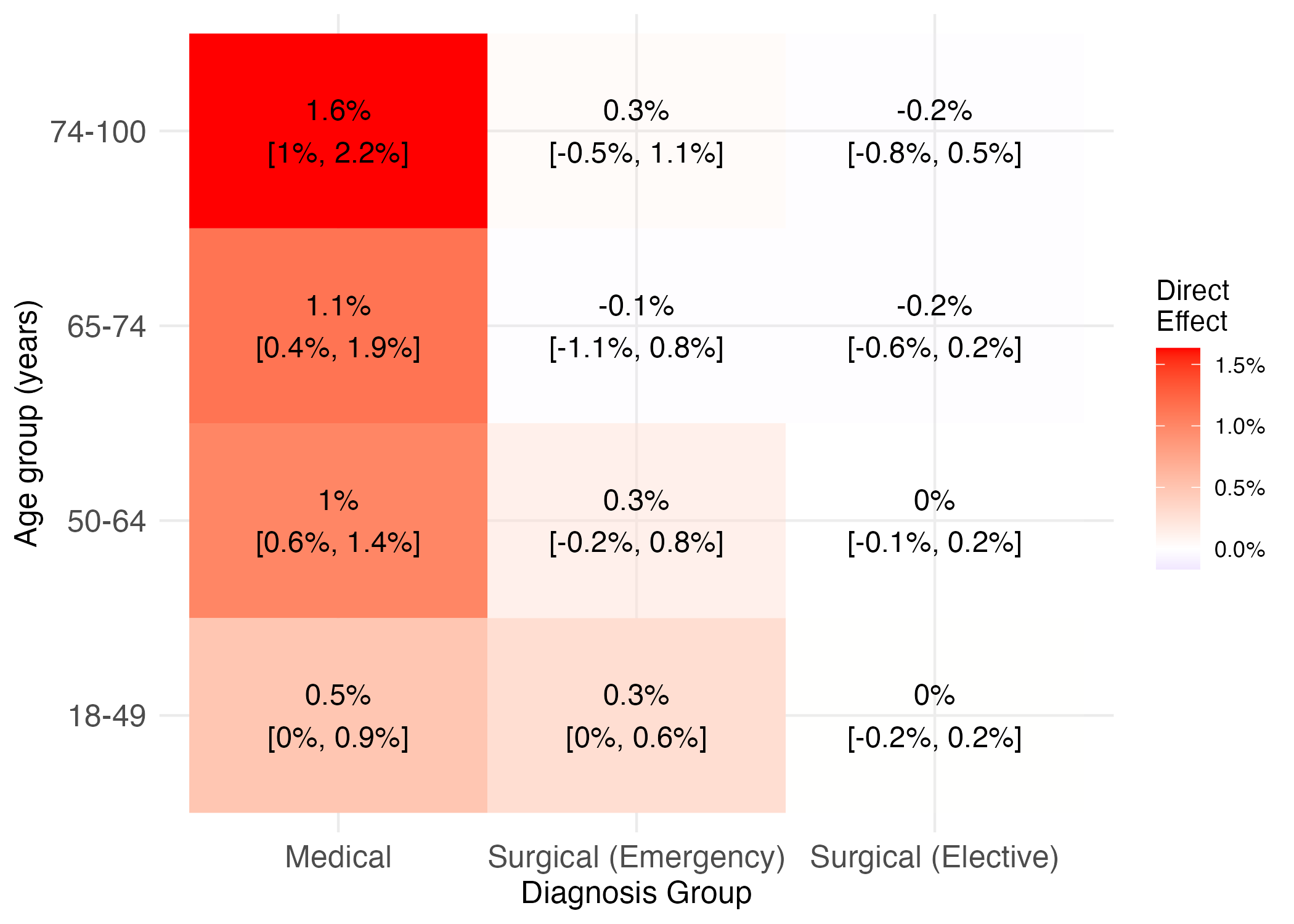}
    };

    \node[box, below=of heterogeneity, align=center, yshift=0.5cm, minimum width = 3.5cm] (brr) {Baseline Risks \\ of ICU Admission};
    \node[box, below=of brr, yshift=0.5cm, align=center, minimum width = 4cm] (radar) {
      IICE Radar \\
      \includegraphics[width=3cm]{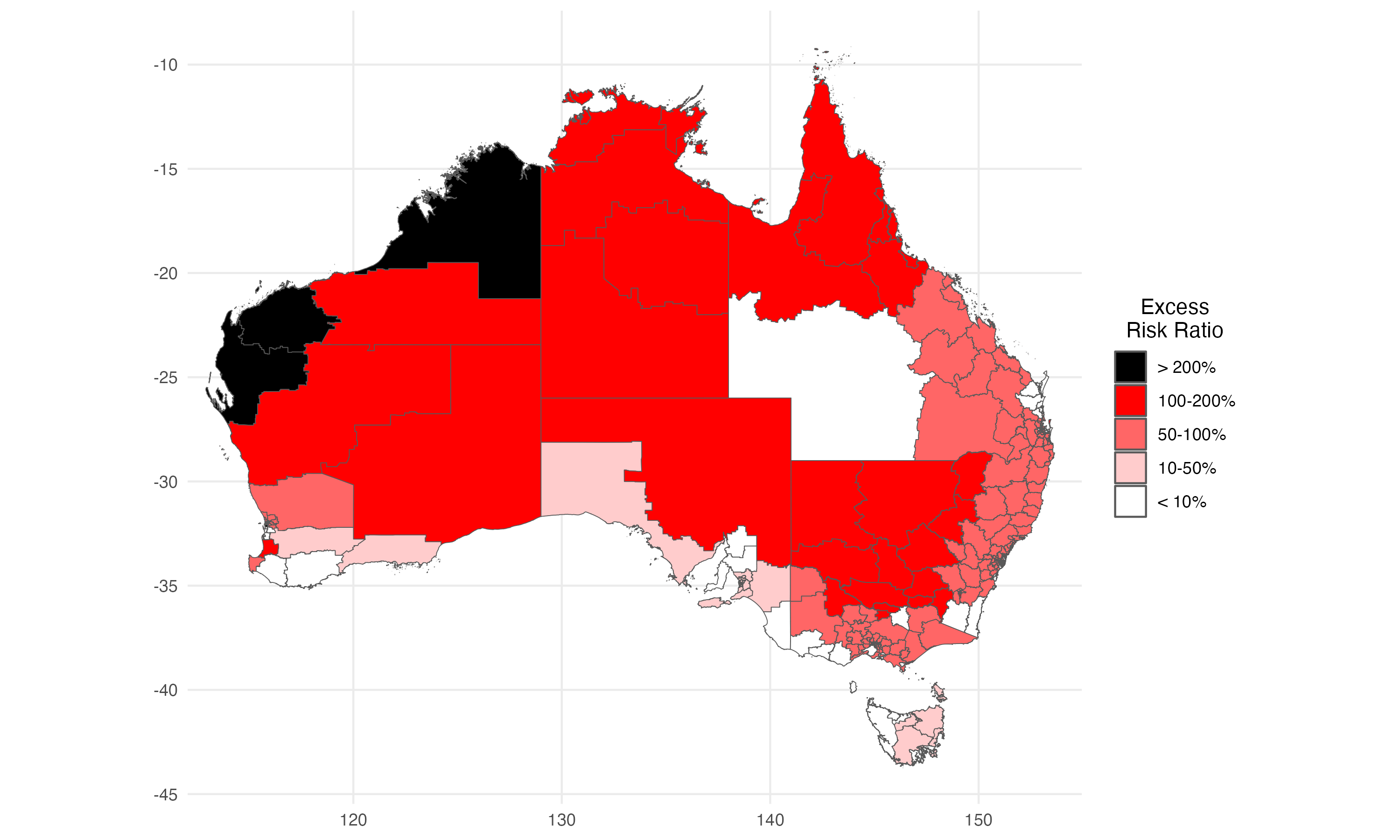}
    };

    % Connections
    \draw[arrow] (centers) -- (anzics);
    \draw[arrow] (us_center) -- (mimic);
    \draw[arrow] ($(datmid) + (-0.0125cm, 0)$) -- (data);
    \draw[arrow] (data) -- (causal);
    \draw[arrow] (model) -- (causal);
    \draw[arrow] (domain) -- (model);
    \draw[arrow] (causal) -- (tv);
    \draw[arrow] (tv) -- (interaction);
    \draw[arrow] (interaction) -- (heterogeneity);
    \draw[arrow] (heterogeneity) -- (brr);
    \draw[arrow] (brr) -- (radar);

\end{tikzpicture}
    }
    %\vspace{1em}
    \caption{Diagram of the performed analysis steps.}
    \label{fig:flowchart}
\end{figure}
\subsection{Data}
The first dataset we analyzed is the Australian and New Zealand (ANZ) Intensive Care Society (ANZICS) Adult Patient Database (APD) \citep{secombe2023thirty}. The ANZICS APD receives submissions from 98\% of ICUs in Australia.
We analyzed all admissions from 181 hospitals across Australia between 2018 and 2024, yielding a cohort of $1,035,890$ patients, of whom $38,736$ ($3.7\%$) were Indigenous. The overall mortality rate was $7.5\%$. 
% and 67\% of ICUs in New Zealand.
The second dataset we analyzed is the Medical Information Mart for Intensive Care (MIMIC-IV) \citep{johnson2020mimic} database, from the Beth Israel Deaconess Medical Center in Boston, Massachusetts. We analyzed all ICU admissions between 2008 and 2019, yielding a cohort of $38,844$ patients, of whom $4,640$ ($11.9\%$) were African-American. The overall mortality rate was $7.6\%$. 
Further patient information for both datasets, including patient filtering steps, and the description of the covariates used in the analysis, can be found in the \nameref{sec:methods} section and Appendix~\ref{appendix:patient-info}.

\subsection{Framework of Causal Fairness Analysis} \label{sec:cfa}
Throughout, we followed the framework of causal fairness analysis described in \citep{plevcko2024causal}. A diagram of the analysis performed in the paper is shown in Fig.~\ref{fig:flowchart}. Using the ricu R-package \citep{bennett2023ricu}, we performed data loading for both datasets, which includes information on race (MIMIC-IV) or ethnicity (ANZICS APD), age, sex, socioeconomic status (SES for short, available in ANZICS APD only), chronic health status, admission diagnosis, illness severity, and in-hospital mortality. 
Based on domain knowledge, we constructed a specific type of causal diagram called the Standard Fairness Model (SFM), in which variables are categorized into four groups (see Fig.~\ref{fig:sfm}). The protected attribute is race/ethnicity (depending on the dataset), labeled $X$. 
The set of confounders, labeled $Z$, includes age, sex, and SES (SES was available only in Australian data). These demographic variables were chosen as confounders since they may be correlated with race/ethnicity but may not necessarily be causally influenced by it.
The set of mediators $W$ includes chronic health status, admission diagnosis, and illness severity. The outcome of interest is in-hospital mortality, labeled $Y$. A comparison of patient characteristics for majority and minority groups for the ANZICS APD and MIMIC-IV cohorts is given in Tbls.~\ref{tab:anzics-pts-tbl} and \ref{tab:mimic-pts-tbl}, respectively.

Our goal in this paper was to explain the observed disparity in mortality between the demographic groups. We started by computing the average differences in mortality rates (also known as the total variation, or TV measure), a commonly used measure of disparity, and found that
\begin{align} \label{eq:aus-tv}
    \textbf{(AU)} \;\; \ex [\text{death} \mid \text{Non-Indigenous}] - \ex [\text{death} \mid \text{Indigenous}] &= -0.4\%,   \\
    \textbf{(US)} \quad\;\;\,\ex [\text{death} \mid \text{White}] - \ex [\text{death} \mid \text{African-American}] &= 0.8\%.  \label{eq:us-tv}
\end{align}
These statistics show that minority patients in Australia were more likely to die after ICU admission, while this finding is reversed in the US data. 
Based on the constructed causal diagram from Fig.~\ref{fig:sfm}, however, we see that the average difference in mortality rates as in Eqs.~\ref{eq:aus-tv}-\ref{eq:us-tv} can arise in three different ways:
\begin{enumerate}[label = (\roman*)]
    \item \textit{confounded/spurious effect:} race/ethnicity may be associated with age, sex, or socioeconomic status, which may influence the mortality risk,
    \item \textit{indirect effect:} race/ethnicity may influence chronic health, admission diagnosis, and illness severity, which have an effect on the mortality risk,
    \item \textit{direct effect:} race/ethnicity may influence the mortality risk, with all other variables kept equal.
\end{enumerate}
Using the framework of causal fairness analysis, we can compute the above three effects, and quantify how much each of the effects contributes to the marginal disparity reported in Eqs.~\ref{eq:aus-tv}-\ref{eq:us-tv}.

\subsection{Decomposing the Disparity} \label{sec:tv-decomp}
\begin{figure}
    \centering
        \begin{subfigure}[b]{0.48\textwidth}
        \centering
        \scalebox{0.9}{
        \begin{tikzpicture}
	 	[>=stealth, rv/.style={thick}, rvc/.style={triangle, draw, thick, minimum size=8mm}, node distance=7mm]
	 	\pgfsetarrows{latex-latex};
	 	\begin{scope}
	 	\node[rv] (c) at (2,1.47) {age};
            \node[rv] (c) at (3,1.5) {sex};
            \node[rv] (c) at (4,1.525) {SES};
	 	%\node[rv] (z2) at (3.75,1.5) {${Z_2}$};
	 	\node[rv] (a) at (0,0) {ethnicity/race};
	 	\node[rv, align=center] (m) at (1.5,-1.5) {admission\\diagnosis};
	 	\node[rv, align=center] (l) at (3.2,-1.5) {chronic\\health};
	 	\node[rv, align=center] (r) at (4.7,-1.5) {illness\\severity};
	 	\node[rv] (y) at (6,0) {mortality};
            \node[rv] (empty) at (0,-2.2) {};
	 	
	 	\node (Zset) [draw,rectangle,minimum width=3.2cm,minimum height=0.8cm,label={[above right]Confounders $Z$}] at (3,1.5) {};
	 	\node (Wset) [draw,rectangle,minimum width=4.8cm,minimum height=1cm, label={[above right]Mediators $W$}] at (3,-1.5) {};
	 	
	 	% edges
	 	\path[->] (a) edge[bend left = 0] (y);
	 	\path[->] (a) edge[bend left = -20] (Wset);
	 	\path[->] (Zset) edge[bend left = 0] (Wset);
	 	\path[->] (Wset) edge[bend left = -20] (y);
	 	\path[->] (Zset) edge[bend left = 20] (y);
	 	
	 	%bidirected
	 	\path[<->,dashed] (a) edge[bend left = 20](Zset);
	 	\end{scope}
	\end{tikzpicture}
        }
        \caption{Standard Fairness Model.}
        \label{fig:sfm}
    \end{subfigure}
    \hfill
    \begin{subfigure}[b]{0.5\textwidth}
        \centering
        \includegraphics[width=\textwidth]{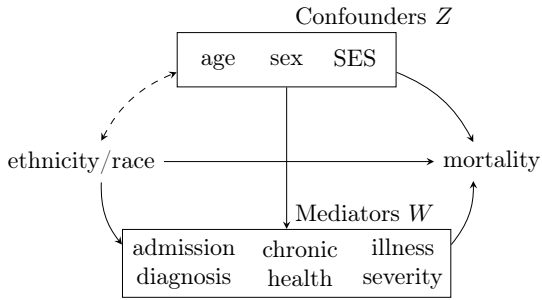}
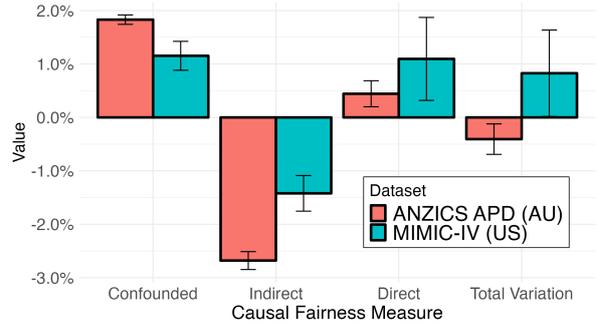
        \caption{Causal decompositions ANZICS APD / MIMIC-IV.}
        \label{fig:tv-decomp}
    \end{subfigure}
    \caption{(a) Standard Fairness Model constructed for the data (b) Decomposition of the marginal disparity in outcome into confounded, indirect, and direct effects, on ANZICS APD and MIMIC-IV datasets.}
    \label{fig:sfm-and-decomp}
\end{figure}

The results quantifying confounded, indirect, and direct effects on both datasets are shown in Fig.~\ref{fig:tv-decomp}.
The TV measure, reported in Eqs.~\ref{eq:aus-tv}-\ref{eq:us-tv} and shown in the last column of Fig.~\ref{fig:tv-decomp}, takes opposite signs on the two datasets. 
However, when applying a causal perspective on the problem, we obtain the following decomposition of the TV measure:
{\fontsize{9.5pt}{11.5pt}\selectfont
\begin{align} \label{eq:aus-tv-decomp}
    \textbf{(AU)} \; \ex [\text{death} \mid \text{Non-Indigenous}] - \ex [\text{death} \mid \text{Indigenous}] &= -0.4\% = \underbrace{1.8\%}_{\text{confounded}} + \underbrace{(-2.7\%)}_{\text{indirect}} + \underbrace{0.5\%}_{\text{direct}},   \\
    \textbf{(US)} \quad\;\;\ex [\text{death} \mid \text{White}] - \ex [\text{death} \mid \text{African-American}] &= \;\;0.8\% \; = \underbrace{1.1\%}_{\text{confounded}} + \underbrace{(-1.3\%)}_{\text{indirect}} + \underbrace{1.0\%}_{\text{direct}}\hspace{-1pt}.  \label{eq:us-tv-decomp}
\end{align}}
\!\!We see that the direct, indirect, and spurious effects (first three columns in Fig.~\ref{fig:tv-decomp}) are in fact equal in sign when applying the causal decomposition. The causal interpretation of the decomposition can be summarized as follows.
First, along the confounded causal pathway, there exists a protective effect for minority patients, transmitted through variables such as age, sex, or SES (accounting for 1.8\% and 1.1\% of the overall variation in the AU and US data, respectively).  
Second, along the indirect causal path, there is a harmful effect on minority patients, indicating that minority patients are more likely to die as a result of the indirect effect (accounting for negative 2.7\% and 1.3\% of the overall variation, respectively). Variables that transmit the indirect effect include chronic health status, admission diagnosis, and the degree of illness severity.
Third, along the direct causal path (when keeping all other variables equal), there is a protective effect of belonging to the minority group, which accounted for 0.5\% of the overall variation in the AU data, and 1.0\% in the US data. 
In other words, for two individuals with comparable characteristics (age, chronic health, illness severity, admission diagnosis) who differ with respect to the protected attribute, the one belonging to the minority group is more likely to survive after ICU admission.
Finally, we emphasize that all the effects studied were statistically significant, indicating that the average differences between groups are in fact a complex interplay of multiple causal mechanisms that transmit change between these demographics.

\begin{figure}[t]
    \centering
    % First row
    \begin{subfigure}[b]{0.3\textwidth}
        \includegraphics[width=\textwidth]{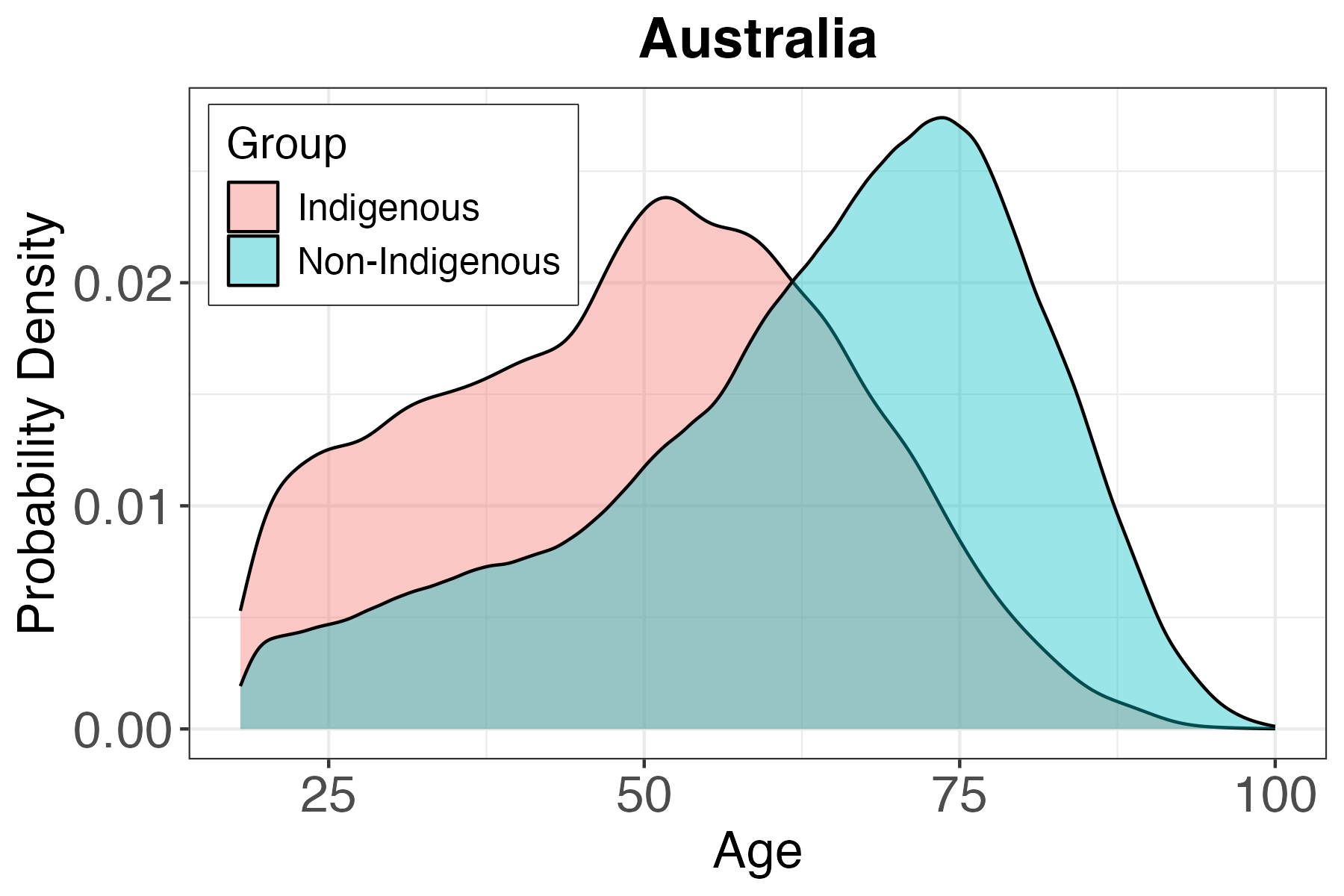}
        \caption{}
        \label{fig:age-aics}
    \end{subfigure}
    \hfill
    \begin{subfigure}[b]{0.3\textwidth}
        \includegraphics[width=\textwidth]{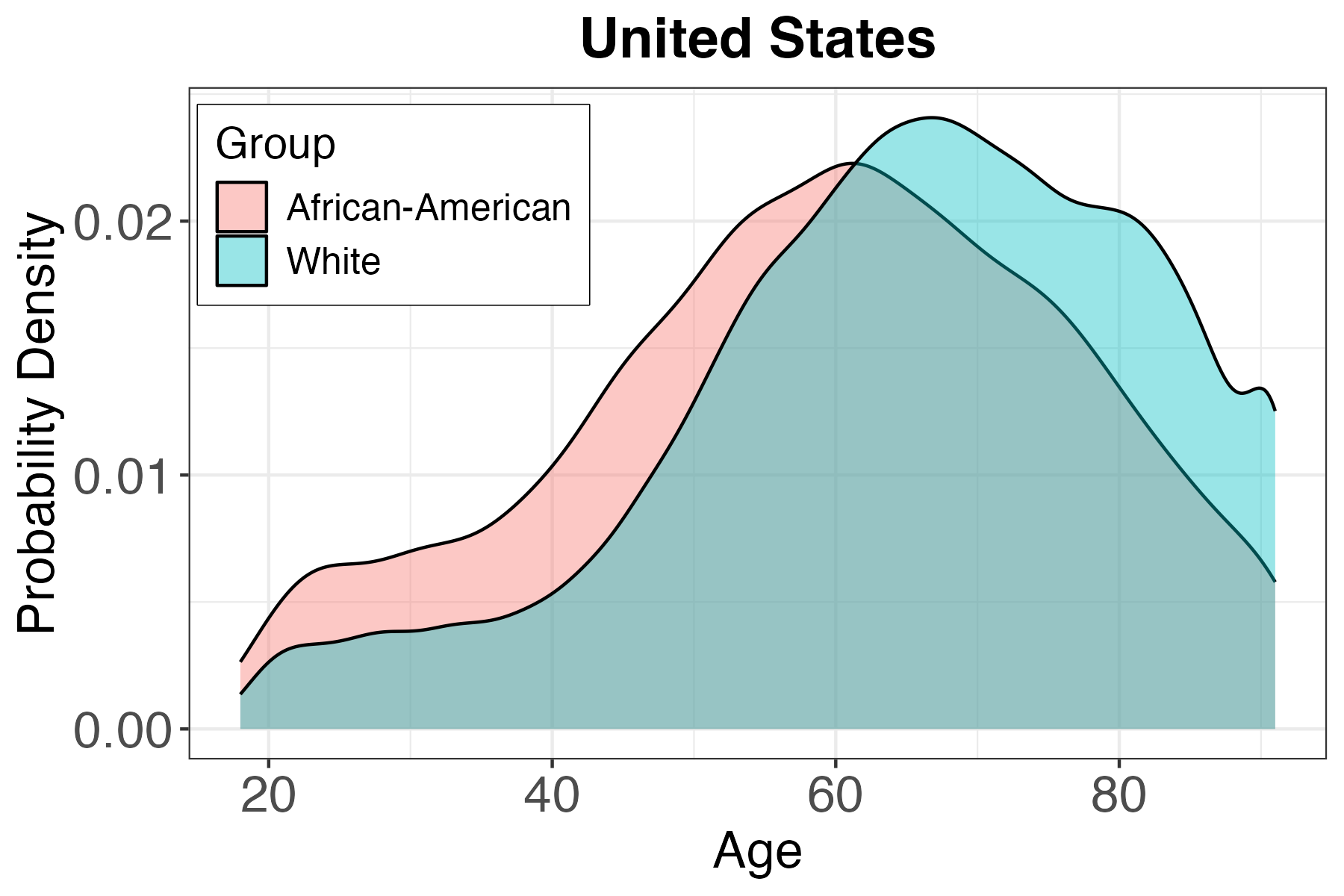}
        \caption{}
        \label{fig:age-miiv}
    \end{subfigure}
    \hfill
    \begin{subfigure}[b]{0.3\textwidth}
        \includegraphics[width=\textwidth]{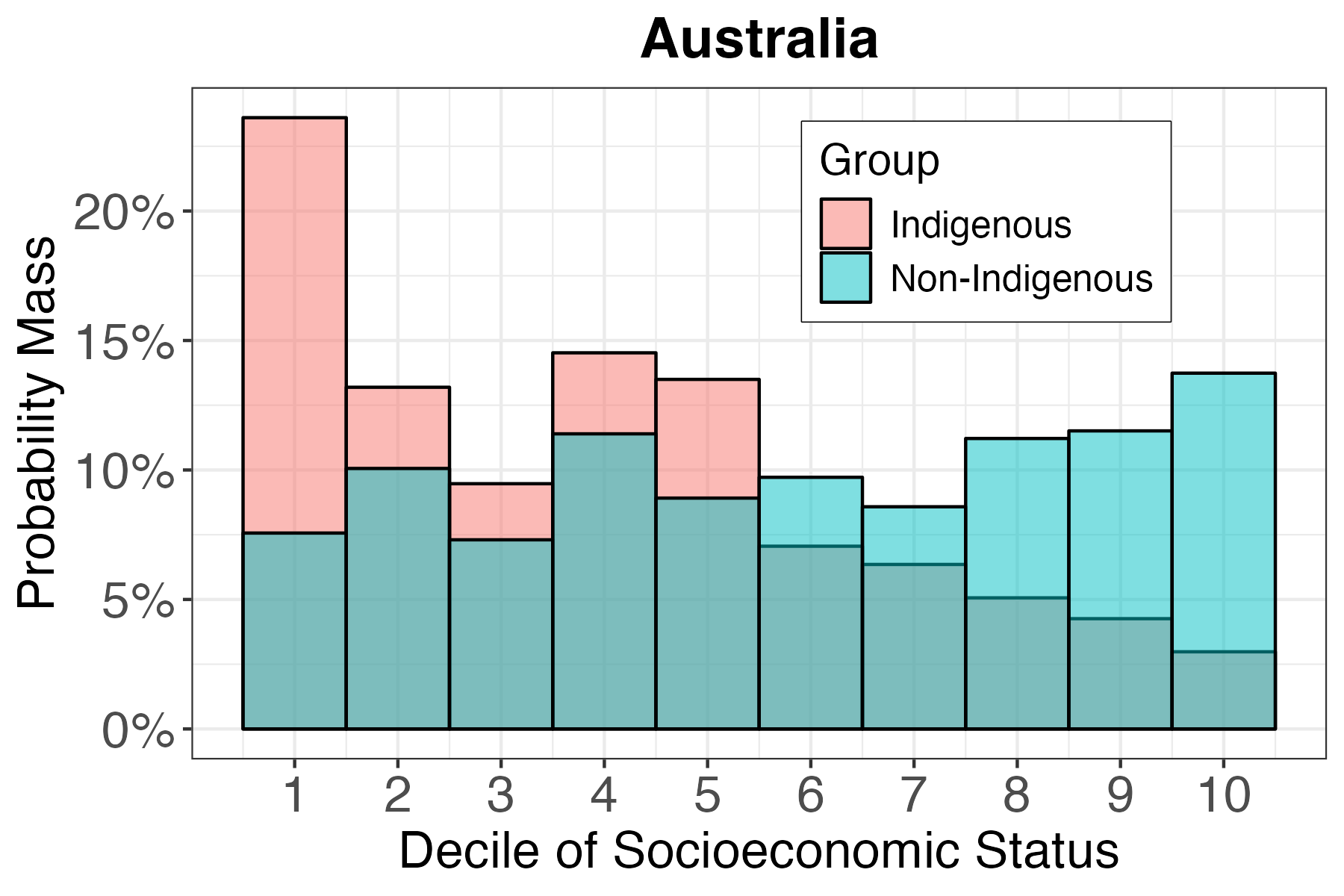}
        \caption{}
        \label{fig:ses-aics}
    \end{subfigure}
    
    % Second row
    \vspace{0.3em}
    \begin{subfigure}[b]{0.3\textwidth}
        \includegraphics[width=\textwidth]{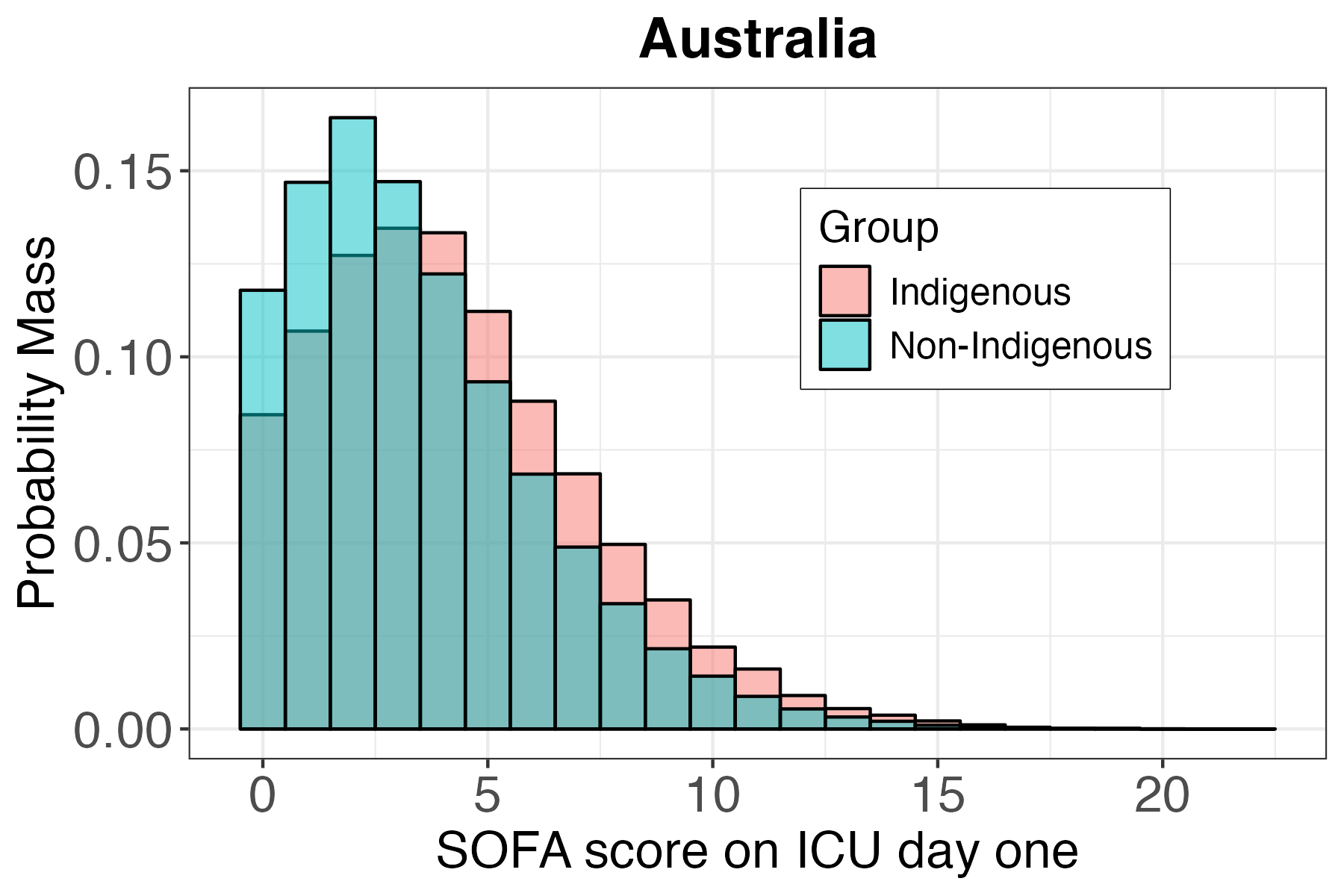}
        \caption{}
        \label{fig:ap3-rod-aics}
    \end{subfigure}
    \hfill
    \begin{subfigure}[b]{0.3\textwidth}
        \includegraphics[width=\textwidth]{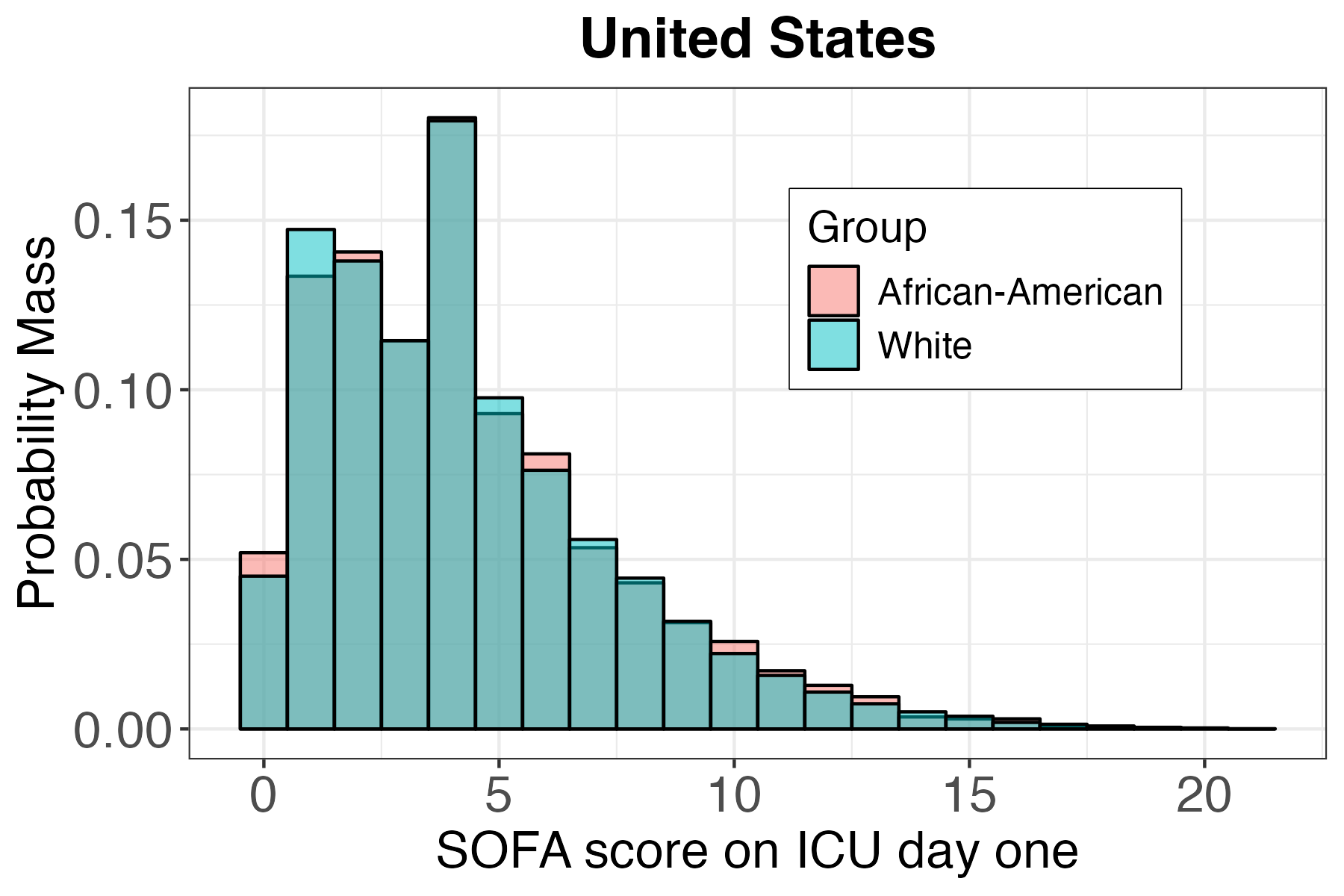}
        \caption{}
        \label{fig:sofa-miiv}
    \end{subfigure}
    \hfill
    \begin{subfigure}[b]{0.3\textwidth}
        \includegraphics[width=\textwidth]{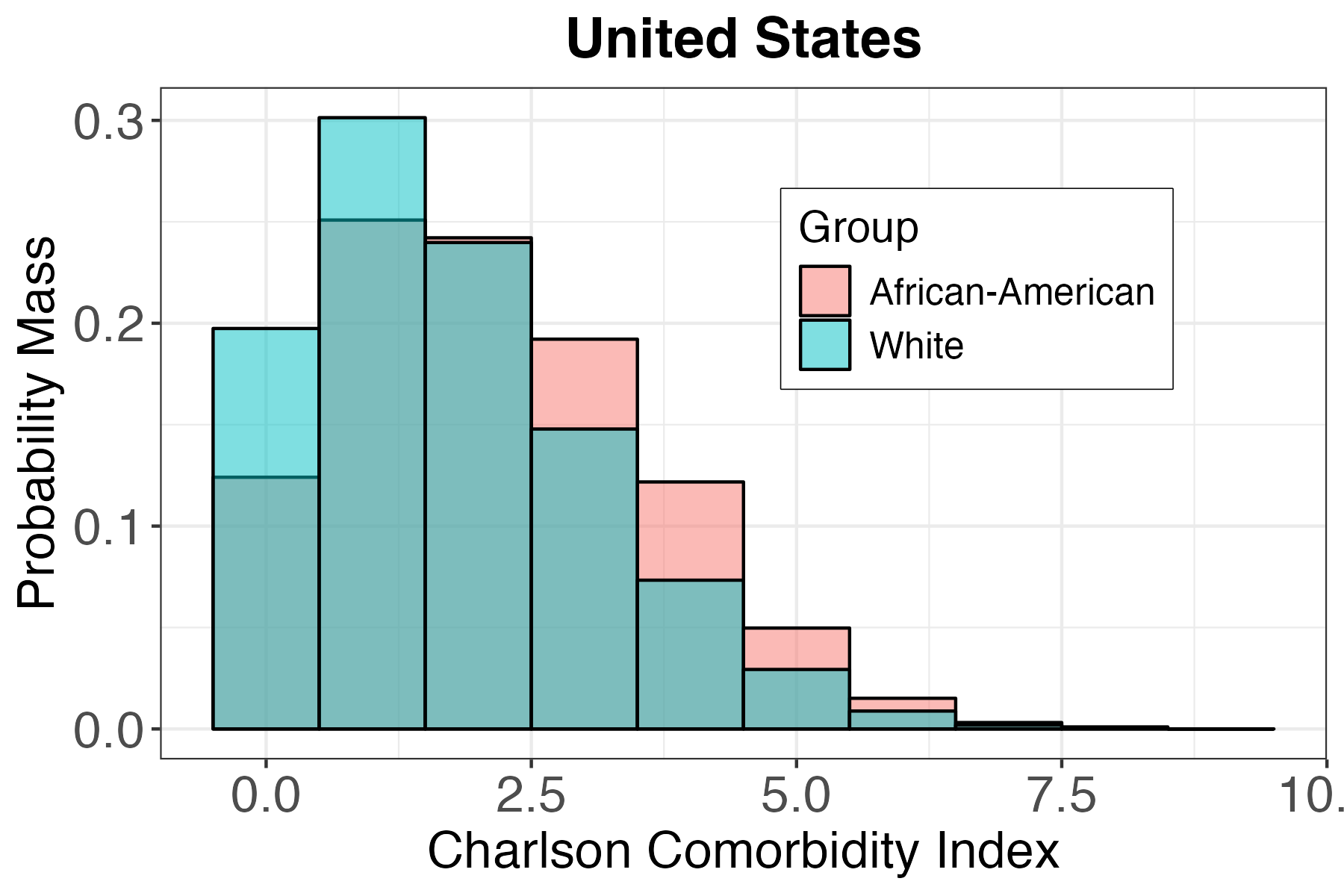}
        \caption{}
        \label{fig:charlson-miiv}
    \end{subfigure}
    \caption{(a) Age in Australian data; (b) Age in US data; (c) Deciles of socioeconomic status in Australian data; (d) SOFA score in Australian data; (e) SOFA score in US data; (f) chronic health status in US data (Charlson comorbidity index).}
    \label{fig:main}
\end{figure}

\noindent\textbf{Explaining the confounded effect.} 
%We first take a closer look at the confounded effect. 
The confounding variables between the protected attribute ethnicity/race and the outcome are age, sex, and SES (see Fig.~\ref{fig:sfm}). 
We therefore analyzed the age/sex distributions of different subpopulations. In both datasets, minority patients were admitted much younger on average compared to their majority counterparts (average difference in age of 13.6 years in Australian, and 6.3 years in US data). 
The age distributions for the two countries, across minority and majority groups, are shown in Figs.~\ref{fig:age-aics}, \ref{fig:age-miiv}. 
Further, we found differences in socioeconomic status in the Australian data (see Fig.~\ref{fig:ses-aics}). Indigenous patients have lower SES on average, which increases the mortality risk.
Since younger age and higher SES reduce the mortality risk, the group-specific differences in age/SES are both relevant for understanding the confounded effect in the first column of Fig.~\ref{fig:tv-decomp}.

\noindent\textbf{Explaining the indirect effect.} The indirect effect, reported in the second column of Fig.~\ref{fig:tv-decomp}, is protective for the majority group. 
This effect is mediated by admission diagnosis, illness severity, and chronic health status. Therefore, we compared group-specific distributions of illness severity after adjusting for age. 
In Australia, minority patients had higher illness severity (Fig.~\ref{fig:ap3-rod-aics}, $p < 0.001$ for difference in means), while the distribution of illness severity in the US data did not show a clear trend (Fig.~\ref{fig:sofa-miiv}, $p = 0.30$ for difference in means).
We also analyzed differential patterns in admission diagnosis after adjusting for age. Minority patients were more likely to be admitted for a medical vs. a surgical reason (both $p < 0.001$) and were more likely to be a non-elective admission (both $p < 0.001$). This mechanism also explains a part of the observed indirect effect, since medical and non-elective admissions carry a greater risk of death.
Furthermore, in the US data, minority patients had a higher average Charlson comorbidity index \citep{charlson1987new} (Fig.~\ref{fig:charlson-miiv}, $p < 0.001$), while in the Australian data, minority patients scored higher on the Clinical Frailty Scale \citep{rockwood2005global} ($p < 0.001$). Therefore, worse pre-morbid health status also contributed to the indirect effect. 
In summary, the indirect effect is explained through a combination of differences in admission pattern, chronic health status, and illness severity.

\noindent\textbf{Explaining the direct effect.} In both datasets, the direct effect was protective for minority patients when all other variables were kept equal. This finding required an extended analysis, which is described next. 
First, we note that possible unobserved confounders for the direct effect of race/ethnicity on outcome include variables such the SES (measured only in AU data) or other unobserved social determinants of health (SDoH). However, the above analysis shows that SES (when available) indicates better status for the majority group. Thus, the addition of SES or SDoH data to the analysis would likely not explain away the protective direct effect for the minority group, but rather make it more pronounced. To better understand direct effects, we took a more granular approach to the quantification of such effects. 

\subsection{Interaction Testing and Heterogeneous Effects} \label{sec:interaction-testing}
\begin{figure}
    \centering
    \begin{subfigure}[b]{0.32\textwidth}
        \centering
        \includegraphics[width=\textwidth]{figures/aics-death-crf-diag_grp-age-majority-TRUE.png}
        \caption{Direct effect heterogeneity on death on ANZICS APD.}
        \label{fig:de-age-diag-aics}
    \end{subfigure}
    \hfill
    \begin{subfigure}[b]{0.32\textwidth}
        \centering
        \includegraphics[width=\textwidth]{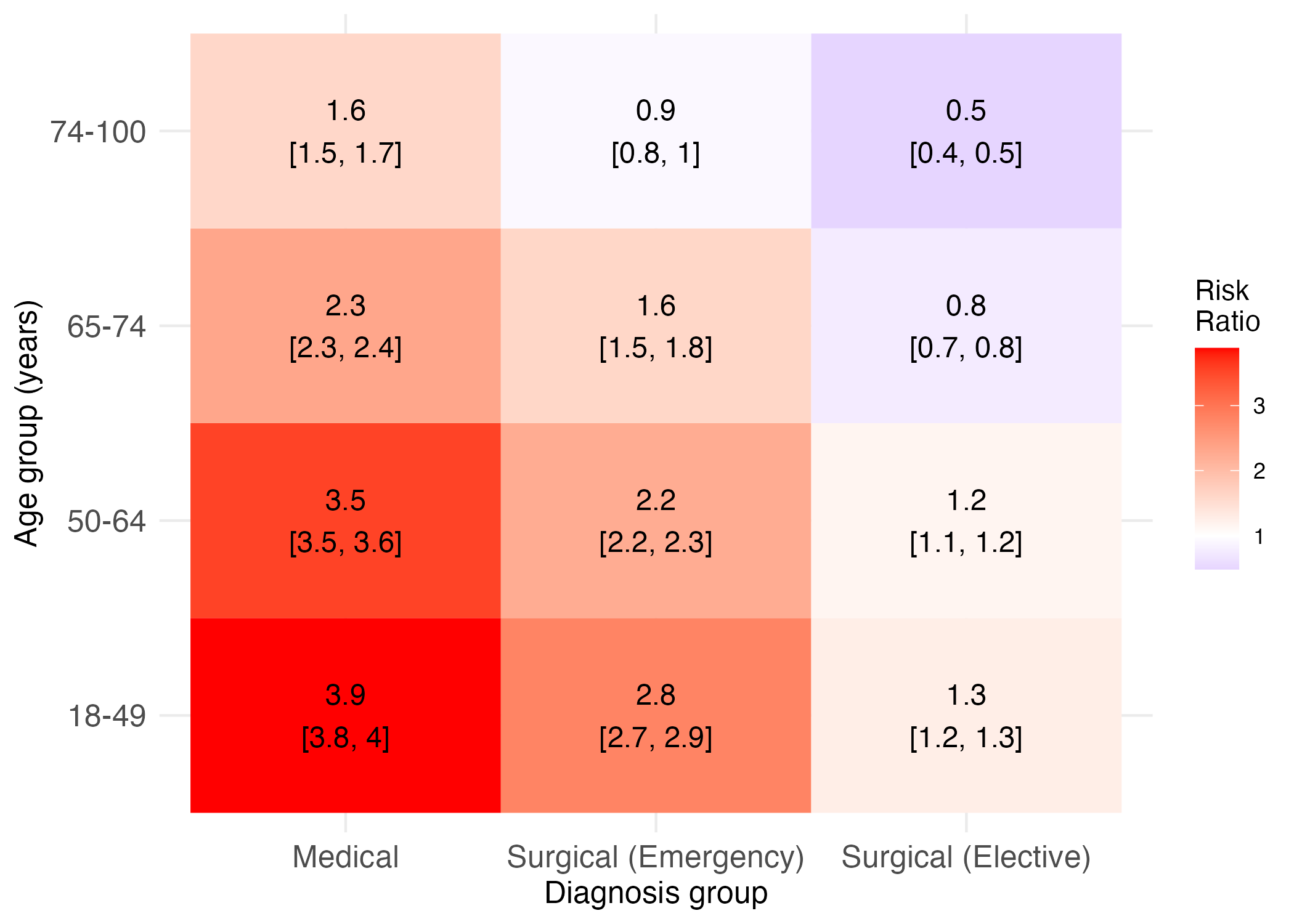}
        \caption{Baseline risk ratio of ICU admission in Australia.}
        \label{fig:brr-age-diag-aics}
    \end{subfigure}
    \hfill
    \begin{subfigure}[b]{0.32\textwidth}
        \centering
        \includegraphics[width=\textwidth]{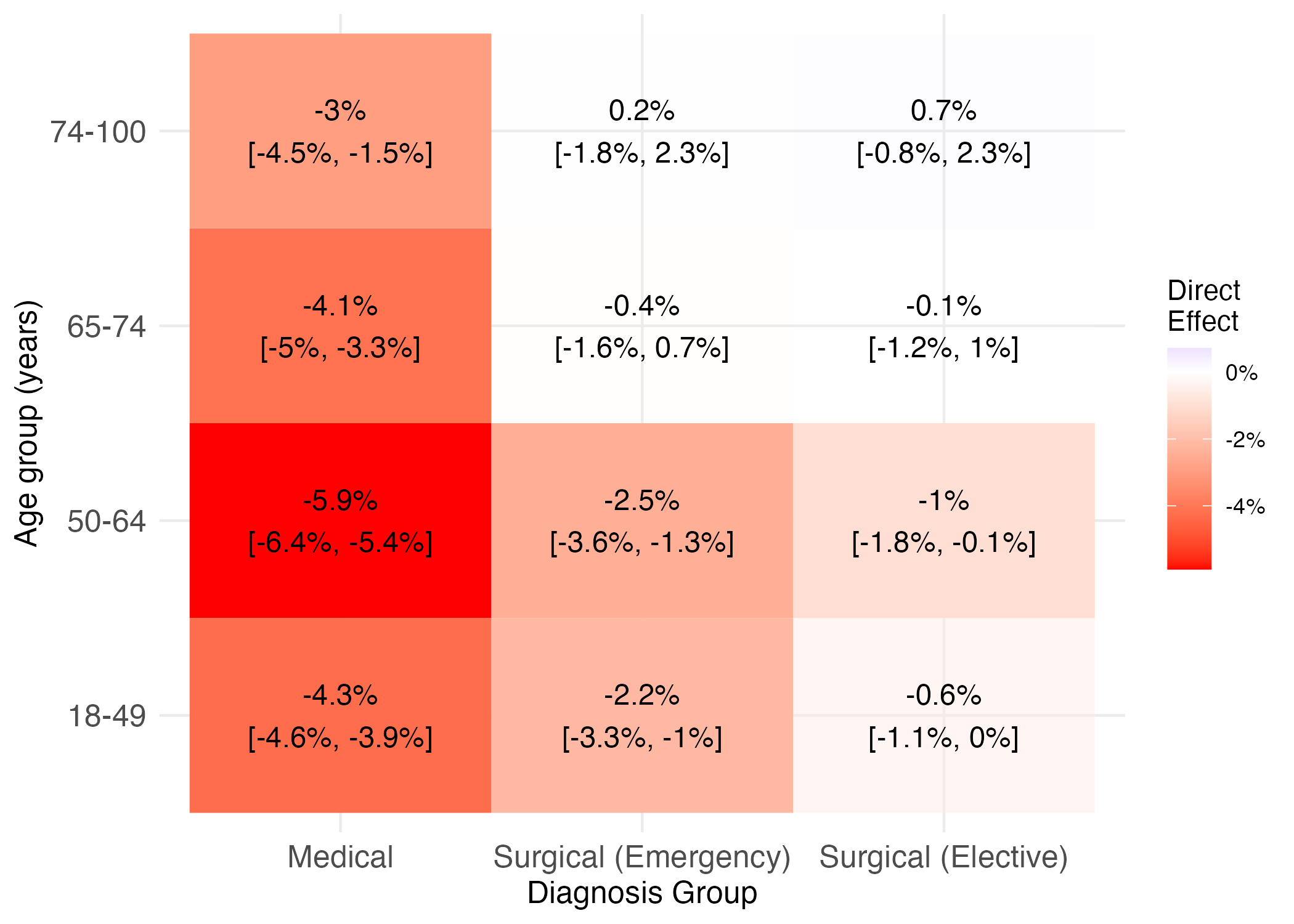}
        \caption{Direct effect heterogeneity on readmission on ANZICS APD.}
        \label{fig:de-age-diag-aics-readmission-aics}
    \end{subfigure}

    \begin{subfigure}[b]{0.32\textwidth}
        \centering
        \includegraphics[width=\textwidth]{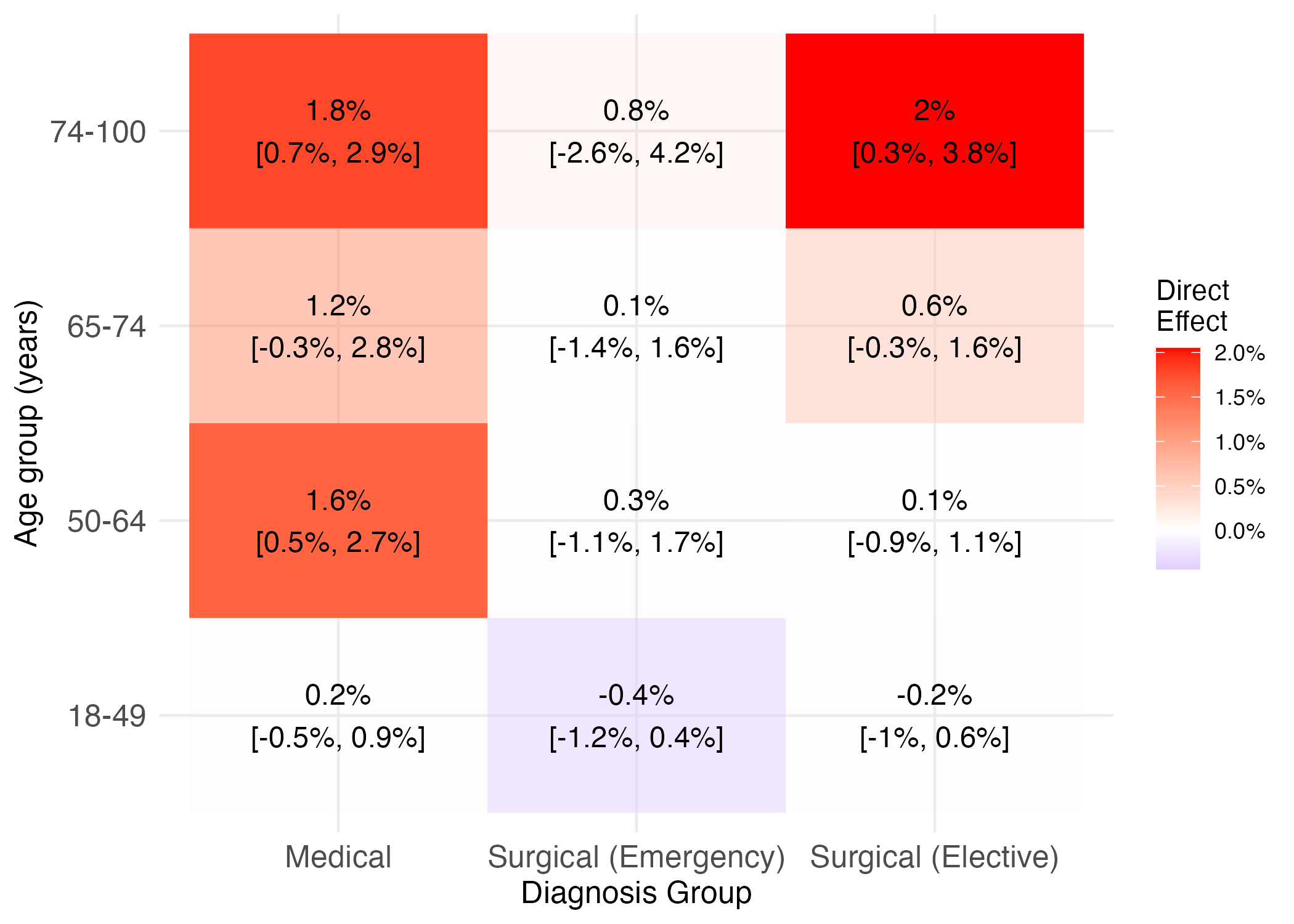}
        \caption{Direct effect heterogeneity on death on MIMIC-IV.}
        \label{fig:de-age-diag-miiv}
    \end{subfigure}
    %\hfill
    \hspace{1in}
    \begin{subfigure}[b]{0.32\textwidth}
        \centering
        \includegraphics[width=\textwidth]{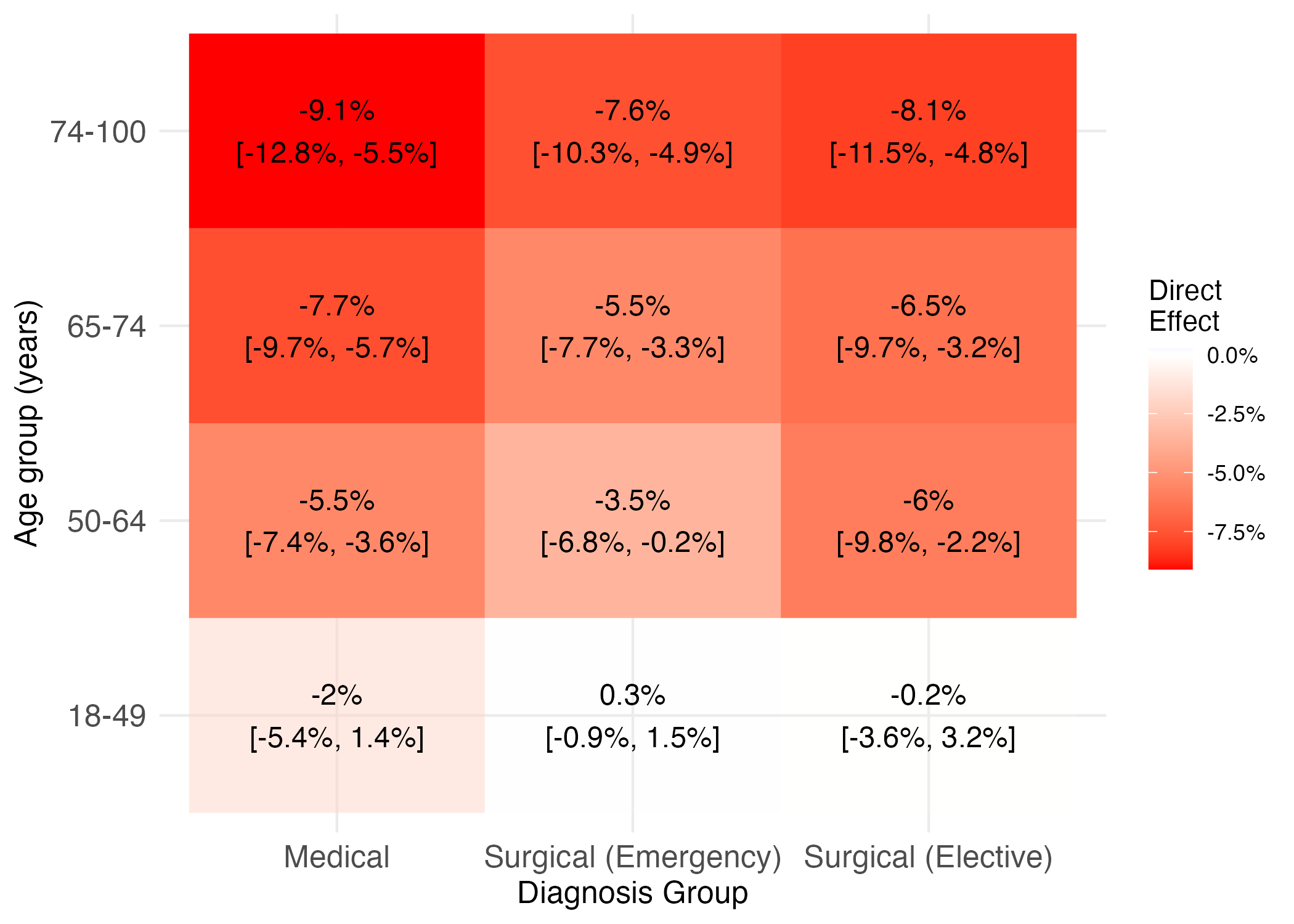}
        \caption{Direct effect heterogeneity on readmission on MIMIC-IV.}
        \label{fig:de-age-diag-aics-readmission-miiv}
    \end{subfigure}
    
    \caption{Heterogeneity according to age group and admission type of (a) direct effect of minority status on death on ANZICS APD; (b) baseline risk ratio of ICU admission in Australia; (c) direct effect of minority status on readmission on ANZICS APD; (d) direct effect of minority status on death on MIMIC-IV; (e) direct effect of minority status on readmission on MIMIC-IV.}
    \label{fig:age-diag-patterns}
\end{figure}

To assess potential interactions between the causal pathways under study, we performed non-parametric interaction testing \citep{plecko2024interaction}. 
This approach provides statistical tests to detect interactions between causal effects along different pathways. 
As discussed, there are three pathways between ethnicity/race ($X$) and mortality ($Y$) in Fig.~\ref{fig:sfm}, namely confounded, indirect, and direct. Using interaction testing, for each pair of pathways, we tested if their interaction was significant.
In both datasets, we identified significant interactions between (i) indirect and confounded pathways and (ii) direct and indirect pathways.
In the ANZICS APD data, we additionally found significant interactions of direct and confounded pathways. The full set of p-values for interaction tests is given in Appendix~\ref{appendix:interaction-testing}. 
The indirect-confounded interaction suggests that the indirect effect along $X\to W \to Y$ is modified by values of the confounders $Z$. 
This modification is studied in Appendix~\ref{appendix:ie-hetero}. 
Similarly, the direct-indirect interaction suggests that the direct effect along $X \to Y$ is modified by values of the mediators $W$. Finally, the direct-confounded interaction indicates that the direct effect is also modified by the confounders $Z$.
We therefore analyzed the heterogeneity of direct effects of race/ethnicity on outcome across admission diagnoses (mediator) and age groups (confounder). 
The results are shown in Fig.~\ref{fig:de-age-diag-aics} for ANZICS APD and Fig.~\ref{fig:de-age-diag-miiv} for MIMIC-IV. The red color indicates a protective direct effect for the minority group. 
The protective direct effect of minority race/ethnicity was present for the group of medical admissions in both datasets.

We considered several hypotheses for explaining the protective direct effect for medical admissions of minority patients. Firstly, this effect could represent a genuine biological difference between the majority and minority groups. However, such biological effect would most likely need to be present for both medical and surgical admissions. Further, the fact that we are observing this effect for two entirely distinct populations (Australian  Indigenous and African-American) that are far apart in terms of ancestral lineages \citep{tishkoff2009genetic, malaspinas2016genomic} makes this explanation less likely. Therefore, this effect is more likely to be a consequence of social differences between groups, since the minority groups on both continents are known to be disadvantaged in their socioeconomic positions. As discussed earlier, however, it is implausible that confounders related to socioeconomic disadvantage would explain away the direct effect that is protective for the minority group.

\subsection{Tipping Over Hypothesis and Population Risks} \label{sec:tipping-over}
We hypothesized that the protective effect of race/ethnicity observed in the data was due to a \textit{tipping over effect} -- minority patients are more likely to require ICU care due to medical complications. The increase in the probability of requiring ICU admission may be a result of worse access to primary health care, since minority groups are less likely to have access to and utilize health care, as reported in previous literature \citep{davy2016access}.
Put differently, for majority patients, medical complications are more likely to be prevented through primary care, and therefore only the more severe cases reach the ICU, causing a selection bias (or left-censoring).
Following this hypothesis, we sought to investigate the baseline risk of ICU admission across different admission groups in the ANZICS APD. This analysis was possible since the ANZICS APD covers 98\% of all ICU admissions in Australia. Data on the age and ethnicity structure of the Australian population were obtained from the Australian Bureau of Statistics \citep{ABS2021, ABS2021Census}. 
After computing the risks of ICU admission, we found that minority patients were 182\% more likely to be admitted to ICU (compared to a majority group patient) for a medical reason, 79\% more likely for emergency surgery, and 14\% less likely for elective surgery (all $p < 0.001$).

We sought to investigate if the risk ratio for ICU admission was related to the strength of the observed protective direct effect for the minority group. 
A possible causal mechanism of the phenomenon may be described as follows. Larger risk ratios for ICU admission indicate that a larger number of patients were admitted for a specific diagnosis. A higher prevalence of a specific diagnosis may decrease the underlying illness severity, since ICU admissions for patients with conditions that may have been prevented at an earlier stage of care are likely to reduce the overall risk in the group. 
To investigate this potential explanation, we assessed the age-adjusted risk ratio of ICU admission across age groups and admission types, defined as:
\begin{align}
    \frac{P(\text{admission } d \mid do(\text{minority}), \text{ age group})}{P(\text{admission } d\mid do(\text{majority}), \text{ age group})}.
\end{align}
The larger the ratio, the more likely minority patients are to be admitted for this diagnosis and age group, compared to majority patients. The risk ratios across age groups and admission types are presented in Fig.~\ref{fig:brr-age-diag-aics}. 
We tested the similarity of the risk ratios in Fig.~\ref{fig:brr-age-diag-aics} with the estimates of the direct effect of minority status on mortality in Fig.~\ref{fig:de-age-diag-aics}. These effects were significantly correlated (Pearson's correlation coefficient $\rho = 0.45$ with 95\% confidence interval $[0.22, 0.68]$). 
Therefore, the pattern of increased risk of ICU admission appeared to be related to the protective effect of minority status on mortality for different age-diagnosis groups.
Logically, however, other unknown factors (such as baseline disease incidence) would provide further insight into this analysis.

\subsection{Readmission Analysis} \label{sec:readmission}
After examining the patterns of protective direct effects on mortality, and baseline risk of ICU admission, we further investigated the risk of ICU readmission. 
Within the cohort of patients who survived their hospital stay, we analyzed whether the patient was readmitted to ICU after being released from the hospital, and this readmission outcome was labeled $R$ (in the Australian data, readmission to any hospital in the database was considered, while in the US data only readmission to the same hospital was available). 
Using the same quantitative approach as in Sec.~\ref{sec:cfa}, and the same causal model in Fig.~\ref{fig:sfm} with readmission $R$ replacing the mortality outcome $Y$, we estimated the direct effect of minority status on readmission (along the direct $X \to R$ pathway).
The values of this direct effect along different age-admission groups are shown in Fig.~\ref{fig:de-age-diag-aics-readmission-aics} for ANZICS APD and in Fig.~\ref{fig:de-age-diag-aics-readmission-miiv} for MIMIC-IV. Here, the red color indicates an increased risk of readmission for minority patients compared to their majority group counterparts. In both datasets, minority patients were more likely to be readmitted, with a stronger pattern in the US data.
In both countries, the increase in readmission had significant similarity with the protective direct effect pattern for mortality ($\rho = 0.74$ with 95\% CI $[0.53, 0.94]$ for Australia, comparing Figs.~\ref{fig:de-age-diag-aics}, \ref{fig:de-age-diag-aics-readmission-aics}; $\rho = 0.81$  with 95\% CI $[0.58, 1]$ for the US, comparing Figs.~\ref{fig:de-age-diag-miiv}, \ref{fig:de-age-diag-aics-readmission-miiv}). 
In Australia, the increase in readmission also had significant similarity with the baseline risk of ICU admission ($\rho = 0.88$, 95\% CI $[0.81, 0.97]$ for comparing patterns in Figs.~\ref{fig:brr-age-diag-aics}, \ref{fig:de-age-diag-aics-readmission-aics}).

\subsection{Indigenous Intensive Care Equity (IICE) Radar} \label{sec:iice-radar}
\begin{figure}
    \centering
    \includegraphics[width=0.9\linewidth]{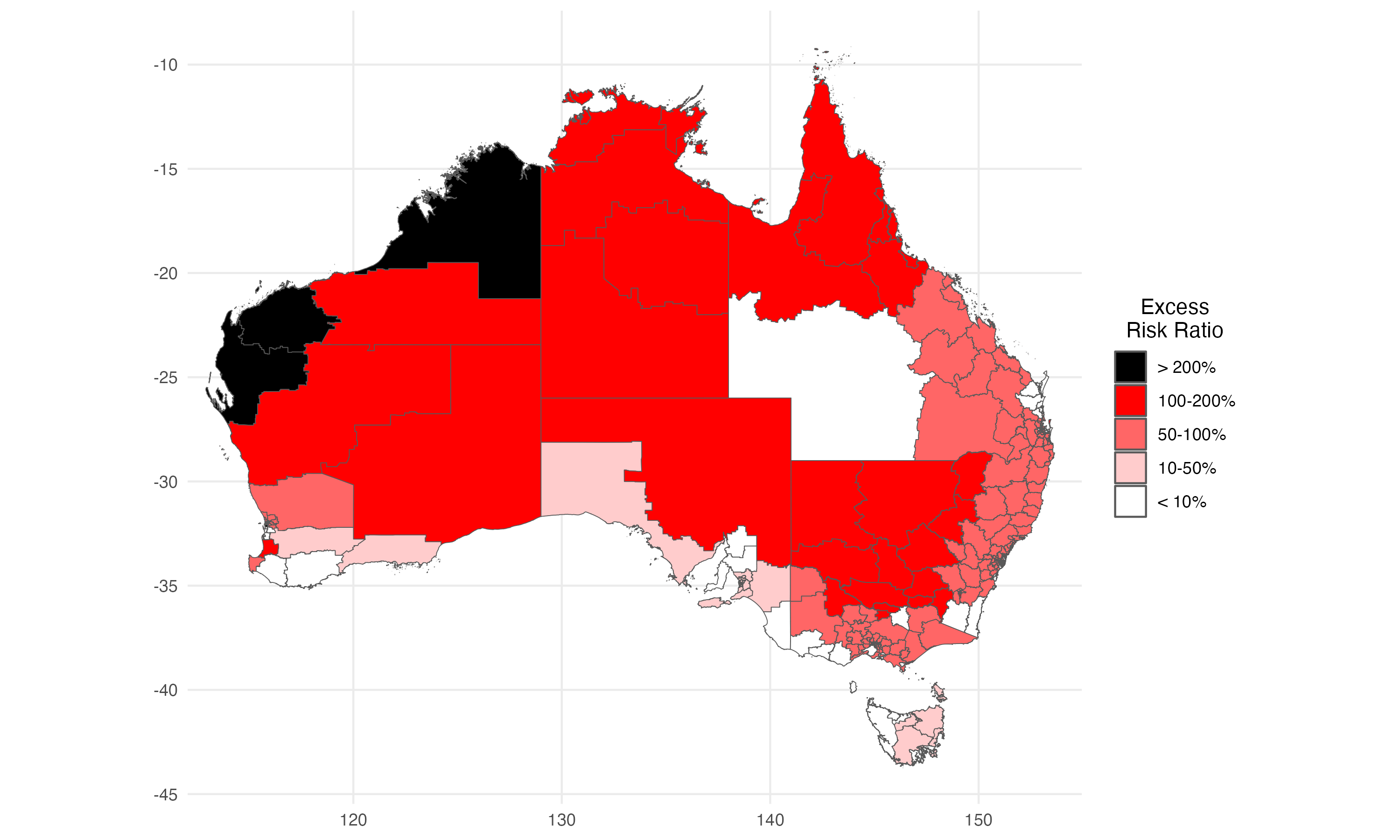}
    \caption{Indigenous Intensive Care Equity (IICE) Radar.}
    \label{fig:iice-radar}
\end{figure}
The fact that Indigenous patients showed improved survival compared to Non-Indigenous patients (with all other variables kept equal) suggests that their actual condition was less severe than the data showed. At the same time, the higher likelihood of Indigenous patients (i) being admitted to the ICU and (ii) being readmitted to the ICU, together with the significant correlation between these patterns and the protective effect on mortality, suggests that these patterns may be jointly explained by limited access to or under-utilization of primary care by the minority group.
Based on this, we hypothesized that the increase in the baseline risk of ICU admission should be investigated as a proxy variable for the under-utilization of primary care by Indigenous patients.
This motivated the construction of the Indigenous Intensive Care Equity (IICE) Radar. From the Australian Bureau of Statistics \citep{ABS2021Census}, we extracted data about age and Indigenous status across different geographical areas of Australia. We then matched areas with the patient information available in the ANZICS APD, to compute the age-adjusted baseline risk ratio of ICU admission, i.e., how much more likely minority patients were to be admitted to ICU depending on the area (our analysis was performed on the level of Statistical Areas 3 used by the Australian Bureau of Statistics for the year 2021). The IICE Radar is shown in Fig.~\ref{fig:iice-radar}, with five risk categories for the different areas. Excess risk ratio of less than 10\% was considered as no risk (white color). Excess RR of 10-50\% was classified as moderate risk (light pink), 50-100\% as substantial (light red), 100-200\% as severe (red), and >200\% as extreme (black). Many of the statistical areas experienced at least moderate risk, implying Indigenous patients were 50\% to a 100\% more likely to be admitted to ICU. Some parts of Western Australia (South West - Perth) recorded excess risk ratios of more than 300\% (minority patients were four times more likely to be admitted), while in other parts (Outback North) the excess risk was up to 900\% (minority patients were ten times more likely to be admitted to ICU). The IICE Radar uncovers substantial differences in risk of ICU admission across different regions, requiring further investigation. The construction of the Radar opens the door to monitoring the increase in admission to ICU, and future studies of access to primary care and its impact on health outcomes of minority groups. 
\section{Methods}\label{sec:methods}

\subsection{Dataset Description and Covariates}
The ANZICS APD contains information on demographics (including age, sex, Indigenous status), biochemical and physiological values during the first 24 hours of the ICU stay, and ICU admission diagnosis (ANZICS modification of the APACHE IV diagnosis list). The information is used for the calculation of severity of illness scores including Sequential Organ Failure Assessment (SOFA), Acute Physiology and Chronic Health Evaluation (APACHE) III scores, APACHE III predicted risk of death, and in-hospital outcomes. 
The covariates extracted for purposes of analysis were:
\begin{itemize}
    \item protected attribute $X$: indigenous status (Indigenous vs. Non-Indigenous),
    \item confounders $Z$: age, sex, and postcode-based socioeconomic status (SES),
    \item mediators $W$: 
    % Clinical Frailty Scale \citep{rockwood2005global} (1-8 scale),
    % count of APACHE III comorbidities (0-7), 
    the APACHE III \citep{knaus1991apache} predicted risk of death, ANZICS modified APACHE-III admission diagnosis (see \href{https://www.anzics.org/wp-content/uploads/2021/03/ANZICS-APD-Dictionary-Version-6.1.pdf}{full list} of considered diagnoses), indicator of whether admission was elective,  
    \item outcome $Y$: in-hospital mortality.
\end{itemize}
SES was not directly available in the data. However, patient postcodes were available, and the postcode information was matched to the Index of Relative Socioeconomic Advantage and Disadvantage (IRSAD) developed by the Australian Bureau of Statistics \citep{abs2021seifa}. This provided information on SES, while $10,911$ ($1.1\%$) patients did not have a recorded postcode. For these patients, a fixed value of -1 was imputed to indicate a missing value. 

The MIMIC-IV dataset is a publicly available resource with comprehensive information on patients admitted to the Beth Israel Deaconess Medical Center (BIDMC), in Boston, Massachusetts. It is sourced from two in-hospital database systems, a custom hospital wide EHR and an ICU specific clinical information system. From the dataset, we extracted the following information:
\begin{itemize}
    \item protected attribute $X$: race (White and African-American),
    \item confounders $Z$: age and sex,
    \item mediators $W$: the Sequential Organ Failure Assessment (SOFA) \citep{vincent1996sofa} score at 24 hours into ICU stay, the Charlson comorbidity index \citep{charlson1987new}, admission diagnosis from one of the 20 categories available in the MIMIC-IV data (see \href{https://mimic.mit.edu/docs/iv/modules/hosp/services/}{full list} of considered diagnoses), indicator of whether admission was elective,  
    \item outcome $Y$: in-hospital mortality.
\end{itemize}
Study flowcharts with patient filtering steps are available in Fig.~\ref{fig:study-flowchart} in Appendix~\ref{appendix:patient-info}. Final cohort numbers were $n=1,035,890$ for ANZICS APD and $n=38,844$ for MIMIC-IV data.

R Statistical Software Version 4.3.2 was used \citep{r2021core} for all the analyses. Data loading was performed using the \text{ricu} R-package \citep{bennett2023ricu}. The README file in our \href{https://github.com/dplecko/RaceMortalityICU}{Github repository} includes instructions for: (i) installing dependencies; (ii) running a demo analysis requiring no data access; (iii) reproducing each figure in the paper; (iv) setting up the MIMIC-IV and ANZICS APD data sources (for which access needs to be obtained).

\subsection{Decomposing the Disparity in Outcome}
We follow the framework of Causal Fairness Analysis described in \citep{plevcko2024causal}, which is based on the language of structural causal models (SCMs) \citep{pearl:2k}. Our approach of analyzing and aggregating findings across multiple data sources is also related to the data-fusion paradigm in the causal inference literature \citep{bareinboim2016causal}.
An SCM is a tuple $\mathcal{M} := \langle V, U, \mathcal{F}, P(u)\rangle$, where $V$, $U$ are sets of endogenous (observable) and exogenous (latent) variables, 
respectively, $\mathcal{F}$ is a set of functions $f_{V_i}$,
one for each $V_i \in V$, where $V_i \gets f_{V_i}(\pa(V_i), U_{V_i})$ for some $\pa(V_i)\subseteq V$ and
$U_{V_i} \subseteq U$. $P(u)$ is a strictly positive probability measure over $U$. Each SCM $\mathcal{M}$ is associated to a causal diagram $\mathcal{G}$ \citep{bareinboim2020on} over the node set $V$ where an edge $V_i \rightarrow V_j$ exists if $V_i$ is an argument of $f_{V_j}$, and a bidirected edge $V_i \bidir V_j$ exists if the corresponding $U_{V_i}, U_{V_j}$ are not independent. An instantiation of the exogenous variables $U = u$ is called a \textit{unit}. By $Y_{x}(u)$ we denote the potential outcome of $Y$ when setting $X=x$ for the unit $u$, which is the solution for $Y(u)$ to the set of equations obtained by evaluating the unit $u$ in the submodel $\mathcal{M}_x$, in which all equations in $\mathcal{F}$ associated with $X$ are replaced by $X = x$. 
Throughout the paper, we use a specific cluster causal diagram $\mathcal{G}_{\text{SFM}}$ known as the standard fairness model (SFM) \citep{plevcko2024causal} over endogenous variables $\{X, Z, W, Y\}$, shown in Fig.~\ref{fig:sfm}, representing the protected attribute, confounders, mediators, and the outcome, respectively.

Our key goal is to decompose the average difference in mortality rates between majority and minority groups, as given in Eq.~\ref{eq:aus-tv}-\ref{eq:us-tv}.
The average difference in mortality, written $\ex[Y \mid X = x_1] - \ex[Y \mid X = x_0]$, where $Y$ is the outcome, $x_1$ the majority group, $x_0$ minority, can be decomposed as \citep{zhang2018fairness, plevcko2024causal}:
\begin{align}
    \ex[Y \mid X = x_1] - \ex[Y \mid X = x_0] &= \underbrace{\ex[Y_{x_1, W_{x_0}} - Y_{x_0} \mid X = x_0]}_{x\text{-DE}_{x_0, x_1}(y \mid x_0)} \quad \text{(direct)} \nonumber \\ 
    &- \underbrace{\ex[Y_{x_1, W_{x_0}} - Y_{x_1} \mid X = x_0]}_{x\text{-IE}_{x_1, x_0}(y \mid x_0)} \quad \text{(indirect)} \label{eq:tv-decomp} \\
    &- \underbrace{\ex[Y_{x_1} \mid X=x_0] - \ex[Y_{x_1} \mid X=x_1]}_{x\text{-CE}_{x_1, x_0}(y)}. \quad \text{(confounded)} \nonumber
\end{align}
The difference $\ex[Y_{x_1, W_{x_0}} - Y_{x_0} \mid X = x_0]$ captures the effect of changing $X$ from minority group $x_0$ to majority group $x_1$ along the direct causal pathway while keeping the mediators at their natural level $W_{x_0}$, averaged across all minority patients (represented by the conditioning $X = x_0$ in the expectation). The difference $\ex[Y_{x_1, W_{x_0}} - Y_{x_1} \mid X = x_0]$ captures the effect of changing $X$ from majority group $x_1$ to minority group $x_0$ along the indirect causal pathway while keeping $X = x_1$ along the direct causal path, average across individuals. For the indirect effect, we considered \textit{the reverse transition} of changing $x_1 \to x_0$ (as opposed to $x_0 \to x_1$ for the direct effect), and this difference was subtracted from the direct effect quantity.   
Finally, the confounded term $\ex[Y_{x_1} \mid X=x_0] - \ex[Y_{x_1} \mid X=x_1]$ compares the effect of setting $X = x_1$ along direct and indirect pathways for the minority group $x_0$ vs. the majority group $x_1$. The marginal difference $\ex[Y \mid X = x_1] - \ex[Y \mid X = x_0]$ is obtained by subtracting the above-described quantifications of indirect and confounded effects from the quantification of the direct effect as shown in Eq.~\ref{eq:tv-decomp}. In Eqs.~\ref{eq:aus-tv-decomp}-\ref{eq:us-tv-decomp} we report the $x$-DE$_{x_0, x_1}(y \mid x_0)$ effect, and the negative of indirect, and confounded effects, $-x$-IE$_{x_1, x_0}(y \mid x_0)$ and $-x$-CE$_{x_1, x_0}(y)$, respectively. This way of reporting the results allows us to represent the TV measure in a simplified way, written as
\begin{align} \label{eq:tv-simplified}
    \text{TV} = \underbrace{x\text{-DE}_{x_0, x_1}(y \mid x_0)}_{\text{direct}} + \underbrace{(-x\text{-IE}_{x_1, x_0}(y \mid x_0))}_{\text{indirect}} + \underbrace{(-x\text{-CE}_{x_1, x_0}(y))}_{\text{confounded}}.
\end{align}
For estimating the effects from finite samples, we derived the efficient influence functions for each term appearing in Eq.~\ref{eq:tv-simplified}, and performed one-step debiasing \citep{kennedy2024semiparametric} to obtain asymptotically normal estimators, for which the uncertainty estimation can be obtained using the normal approximation. Sensitivity analysis for the impact of data missingness on the inference of results is described in Appendix~\ref{appendix:miss-sens}, while in Appendix~\ref{appendix:overlap} we perform a sensitivity analysis to investigate the effect of overlap on inference.

\subsection{Interaction Testing and Heterogeneous Effects}
When decomposing the marginal disparity into direct, indirect, and confounded effects, a transition $x_0 \to x_1$ along the direct effect is considered, while the reverse transitions $x_1 \to x_0$ along the indirect and confounded effects are subtracted from it (see Eq.~\ref{eq:tv-decomp}). The reason why a reverse transition needs to be considered is the possible existence of interactions, as noted by \citep{vanderweele2015explanation}. Following \citep{plecko2024interaction}, an absence of an interaction would imply that the structural mechanism $f_y$ of $Y$ can be written as:
\begin{align} \label{eq:non-ia}
   f_y(x, z, w, u_y) = f^{(1)}_y (x, u_y) + f^{(2)}_y (w, u_y) + f^{(3)}_y (z, u_y),
\end{align}
where $f_y^{(1)}, f_y^{(2)}$, and $f_y^{(3)}$ are the structural functions corresponding to effects of covariates $X$, $W$, and $Z$, respectively. 
For binary outcomes, the absence of interactions is considered on the risk-scale, meaning that the structural mechanism $f_y$ which returns a binary value is replaced by $p_y(x, z, w) = P(f_y(x, z, w, U_y) = 1)$. 
%%could be considered on a different scale (such as the risk ratio or the odds ratio scales). 
For the binary case, the absence of any interactions would be written as this would be given by the condition:
\begin{align} \label{eq:non-ia-risk}
    p_y(x, z, w) := P(f_y(x, z, w, U_y) = 1) = p^{(1)}_y (x) + p^{(2)}_y (w) + p^{(3)}_y (z),
\end{align}
where $p^{(i)}$ are different functions. In case of no interactions (Eq.~\ref{eq:non-ia-risk}), we have the following implications:
\begin{align} \label{eq:de-sym}
    x\text{-DE}_{x_0, x_1}(y \mid x_0) &= x\text{-DE}_{x_1, x_0}(y \mid x_0) \\
    x\text{-IE}_{x_0, x_1}(y \mid x_0) &= x\text{-IE}_{x_1, x_0}(y \mid x_0) \\
    x\text{-CE}_{x_0, x_1}(y) &= x\text{-CE}_{x_1, x_0}(y) \label{eq:se-sym}
\end{align}
% On the risk ratio scale, the absence of interactions, as in Eq.~\ref{eq:non-ia-rr}, implies:
% \begin{align}
%     \frac{\ex[Y_{x_1, W_{x_0}} \mid X = x_0]}{\ex[Y_{x_0} \mid X = x_0]} = \frac{\ex[Y_{x_1} \mid X = x_0]}{\ex[Y_{x_0, W_{x_1}} \mid X = x_0]},
% \end{align}
% for the direct effect, and analogously for the indirect and confounded effects. 
We follow the approach of \citep{plecko2024interaction} as described above and perform non-parametric interaction testing by performing a hypothesis test for each of the Eqs.~\ref{eq:de-sym}-\ref{eq:se-sym}. %(and analogously for the risk ratio scale).
The obtained p-values for each interaction are provided in Appendix~\ref{appendix:interaction-testing}. 
For quantifying heterogeneous effects, we considered four different age groups $C_1, \dots, C_4$, corresponding to 18-54, 55-64, 65-74, and 75-100 years of age. These groups were chosen as approximate quartiles of age in the data. 
For admission categories, we considered three admission types, namely medical admissions $d_{med}$, emergency surgery admissions $d_{ems}$, and elective surgery admissions $d_{els}$.
For investigating heterogeneity of direct effects, we computed the quantity 
\begin{align} \label{eq:zw-de}
    \text{DE}_{x_0, x_1}(y \mid C_i, d, x_0) = \ex[Y_{x_1,W_{x_0}} - Y_{x_0, W_{x_0}} \mid \text{age} \in C_i, \text{ admission}=d, X = x_0]
\end{align}
for different values of $C_i, d$ (reported in Fig.~\ref{fig:age-diag-patterns}).
For heterogeneity of indirect effects, we investigated the quantity
\begin{align}
    -\text{IE}_{x_1, x_0}(y \mid \text{age} \in C_i, x_0) = \ex[Y_{x_1, W_{x_1}} - Y_{x_1, W_{x_0}} \mid \text{age} \in C_i, X = x_0]
\end{align}
for different values of $C_i$ (see Appendix~\ref{appendix:ie-hetero}). The effects $\text{DE}_{x_0, x_1}(y \mid C_i, d, x_0)$ and $\text{IE}_{x_1, x_0}(y \mid C_i, x_0)$ were estimated using causal forests \citep{wager2018estimation}, while the uncertainty estimates were obtained using bootstrap.

\subsection{Tipping Over Hypothesis and Population Risks}
The Australian Bureau of Statistics provides census-based values of population in each age group for years $2016, 2021$ \citep{ABS2021Census}. 
Based on this data, we performed piecewise linear interpolation in each age group for both the overall and minority populations to determine the population counts between 2018 and 2024.  
After performing the imputation, we were able to compute the number of minority and majority persons who are possibly at risk in each year $t$ and age group $a$, denoted as $N(x_0, a, t), N(x_1, a, t)$, respectively. The number of minority and majority patients admitted in year $t$ and age group $a$ to the ICU for diagnosis group $d$ was labeled $n(x_0, a, d, t), n(x_1, a, d, t)$, respectively. The indicator of whether a person is admitted is labeled with $I$. The probability (or risk) of admission for a diagnosis $D = d$, in age group $a$, demographic group $x$, and year $t$, denoted by $P(I = 1 \wedge D=d \mid \text{age} = a, \text{year} = t, X = x)$, is given by:
\begin{align}
    P(I = 1 \wedge D=d \mid \text{age} = a, \text{year} = t, X = x) &= \frac{n(x, a, d, t)}{N(x, a, t)}.
\end{align}
For any conditioning event $E$, one can consider the $E$-specific risk, defined as
\begin{align}
    P(I = 1 \wedge D=d \mid E, do(X = x)) = \sum_{a, t} P(I = 1 \wedge D=d \mid \text{age} = a, \text{year} = t, X = x) P(a, t \mid E)
\end{align}
where $P(a, t \mid E)$ is defined as
\begin{align}
    P(a, t \mid E) = \frac{\sum_{a,t \in E} N(x_0, a, t) + N(x_1, a, t)}{\sum_{a, t} N(x_0, a, t) + N(x_1, a, t)}.
\end{align}
Then, to obtain the effect of minority status on the risk of ICU admission for any conditioning event $E$, we can compute the $E$-specific risk ratio:
\begin{align}
    \text{RR}(d | E) = \frac{P(I = 1 \wedge D=d \mid E, do(X = x_0))}{P(I = 1 \wedge D=d \mid E, do(X = x_1))}.
\end{align}
In Fig.~\ref{fig:brr-age-diag-aics} we reported RR($d \mid \text{age} \in C_i, D = d$) for medical diagnoses $d_{med}$, emergency surgery diagnoses $d_{ems}$, elective surgery diagnoses $d_{els}$, and across age groups $C_i$ (18-49, 50-64, 65-74, 75-100 years old). For quantifying the correlation of different heatmaps in Fig.~\ref{fig:age-diag-patterns}, we used Pearson's correlation coefficient. Consider a heatmap $H$ consisting of $12$ entries $H_{ij}$, with $i \in \{18\text{-}49, 50\text{-}64, 65\text{-}74, 75\text{-}100 \text{ years}\}$ and $j \in \{\text{Medical}, \text{ Surgery (Emergency)}, \text{ Surgery (Elective)}\}$. For two heatmaps $H, L$, the Pearson's correlation coefficient is defined as
\begin{align}
    \rho(H, L) = \frac{\sum_{i,j} (H_{ij} - \bar H)(L_{ij} - \bar L)}{\sqrt{\sum_{i,j} (H_{ij} - \bar H)^2}\sqrt{\sum_{i,j} (L_{ij} - \bar L)^2}},
\end{align}
where $\bar H := \frac{1}{12}\sum_{i,j} H_{ij}, \bar L := \frac{1}{12}\sum_{i,j} L_{ij}.$ 
When computing correlations for heatmaps of the readmission pattern (Figs.~\ref{fig:de-age-diag-aics-readmission-aics}, \ref{fig:de-age-diag-aics-readmission-miiv}), we considered the negative $-H$ heatmap, since negative entries $-H_{ij}$ measure the increase in the risk of ICU readmission for the minority group.   
Confidence intervals for $\rho(H, L)$ are obtained by taking bootstrap samples $\mathcal{D}^b$ of the original data, and repeatedly estimating $H^{b}, L^{b}$, and $\rho(H^{b}, L^{b})$, based on $\mathcal{D}^b$. The confidence interval for $\rho(H, L)$ is then constructed based on the values $\rho(H^{b}, L^{b})$ across the bootstrap draws, using a Gaussian approximation. 
\section{Discussion}\label{sec:discussion}
In this paper, we introduced a systematic approach for analyzing health disparities, based on the tools of causal fairness analysis \citep{plevcko2024causal}. The framework was illustrated through an analysis of racial/ethnic disparities in ICU outcomes between minority and majority groups in Australia and the United States (see Fig.~\ref{fig:flowchart}). 
Our investigation demonstrated that commonly used, statistical measures to quantify disparity are insufficient for investigating health equity. 
In the Australian cohort, the minority group had a higher average mortality rate than the majority group, while this was reversed in the US cohort. 
However, when taking a more fine-grained, causal perspective on data analysis, the studied effects (direct, indirect, spurious) had the same signs and were consistent across populations. 

Using a causal lens also allowed us to investigate different possible pathways that may transmit change between minority status and the outcome of interest. When considering confounded effects, we found that minority race/ethnicity was associated with lower age at admission in both countries \citep{martin2003epidemiology}, which in turn reduced the risk of mortality. 
This effect is most likely driven by increased chronic health problems, and/or lack of access to primary care for the minority groups. 
For the indirect effects, we found that minority groups had worse chronic health, were more likely to be admitted for urgent (non-elective) and medical reasons, and had higher illness severity levels (after removing the age effect). 
All of these factors are known to increase the mortality risk. 
At the same time, surprisingly, we found that along the direct effect, minority patients showed better survival, when all other variables were held constant (\textit{ceteris paribus}).
A further analysis of effect interactions showed that this direct effect was heterogeneous across admission types. For medical admissions, there was a pronounced protective direct effect for the minority group, while this was not the case for surgical admissions.
We hypothesized that the protective direct effect of race/ethnicity, which applied predominantly to medical admissions, may be due to the fact that minority patients have poor access to primary healthcare, as reported in previous literature \citep{davy2016access}. 
The causal mechanism for such a protective effect would be that worse access to primary care results in a higher prevalence of ICU admission for a specific diagnosis, and that such increased prevalence reduces the overall risk of death in that group, since increased prevalence also implies that less severe and possibly preventable cases reach the ICU.
A related phenomenon of increased prevalence and reduced mortality has been observed in the literature on sex-related ICU disparities \citep{modra2022sex}.
%Our findings showed that the risk of ICU admission was especially increased for the set of medical diagnoses. 
In this context, we analyzed the baseline risk of ICU admission in Australia. We showed that the increase in risk for the minority group was highest in the group of medical admissions (compared to the baseline risk for the majority group), and significantly higher than for surgical admissions.
Such increase in risk constitutes a selection bias (or left-censoring of majority group patients), which may explain the observed direct effect. 
Accordingly, we demonstrated that the pattern of increase in the risk of ICU admission was statistically related to the strength of the protective direct effect observed for different diagnostic and age groups. 
In a separate analysis of readmissions, we found that minority patients were in fact more likely to be readmitted \citep{soto2013healthcare}, and this effect was also strongest within the group of medical admissions. 
The fact that Indigenous patients showed increased risk of admission, improved survival, and a higher chance of readmission, along with the significant similarity of these patterns across age-admission groups, supports the interpretation that under-utilization of primary health care may be one of the causes of our findings.
%serves as a protective net for the majority group, and filters out admissions that are more easily treated at an earlier stage of a healthcare system.
This motivated us to construct the Indigenous Intensive Care Equity (IICE) Radar, which monitors the increase in risk of ICU admission for Indigenous patients across different geographical areas. The increase in risk of ICU admission is a hypothesized proxy of lack of access and reduced utilization of primary care. The construction of the Radar opens the door to important future studies with possibly significant public health implications.

An important finding of our study is that, when considering direct, indirect, and spurious effects, racial and ethnic disparities in ICU outcome are a consequence of differences that happen prior to the time the patient enters the ICU, with key factors being (i) worse chronic health and lower age at admission; (ii) higher risk of a non-elective, urgent admission; (iii) worse access to primary care for earlier treatment of preventable ICU admissions.

Another important takeaway of our study is that the ``sign'' of a causal effect estimate is not a definitive way to conclude whether a protected group is discriminated against. For instance, the confounded association of minority race/ethnicity with lower age at admission, resulted in reduced mortality for the minority group. Similarly, along the direct effect, we also observed a protective effect of minority race/ethnicity. However, upon interpreting these effects, we noted that they are most likely a consequence of socioeconomic disadvantage and worse access to healthcare. Therefore, critical care needs to be interpreted within a multi-layered system of healthcare, and the filtering and selection bias of populations that require critical care has direct implications on the health disparities observed in ICU outcomes.

A major strength of our work is the size and the heterogeneity of the data we studied. We studied almost 1.1 million ICU admissions across two countries on different continents, and our findings were robust and consistent across these countries. 
A second strength is the use of a systematic framework for analyzing health inequities, which allows us to both decompose the observed disparity in average mortality rates into its constitutive causal elements, and also allows us to investigate whether there are significant interactions between different causal pathways. Thirdly, we were able to establish hypotheses that explained why differences along direct, indirect, and spurious pathways occur, and we tested a tipping-over hypothesis to explain the protective direct effect for minority patients.
We further connected the quantification of the direct effects with external data on the baseline risk of ICU admission, adding further evidence to support our explanation of the direct effects, and constructed a monitoring tool that may provide the basis for important public health policy in the future. 

In terms of limitations, we acknowledge the observational nature of our study, and note that some relevant confounders such as socioeconomic status were only partially included in our analyses (missing in US data, available in Australian data). Other unobserved social determinants of health (SDoH) should also be included in future studies. Still, we were able to demonstrate that tools of causal inference may help uncover important patterns in health equity investigations. Furthermore, as elaborated in the main text, such missing confounders would likely not explain away the effects we studied, but rather make the protective direct effect for minority patients even more pronounced. 
Secondly, we note that our analysis of baseline risk of ICU admission did not include the overall prevalence of different diseases in the population but focused on rates of admission to ICU according to different diagnoses. Disparities in chronic health are also inherently related to limited access to primary care, since not accessing primary care may result in worse management of chronic health issues \citep{rothman2003chronic}.
Future studies on minority health disparities in ICU outcomes should attempt to further disentangle these different contributing factors and their impact on ICU admission risk.
%However, our findings on the increased admission, improved survival, and increased readmission do seem to support the hypothesis that the observed differences are related to under-utilization of primary care.

\section{Conclusion}
In a large study of almost 1.1 million ICU admissions in Australia and the United States, we used the framework of Causal Fairness Analysis to investigate whether tools of causal reasoning can help to disentangle the mechanisms underlying disparities in mortality according to race and ethnicity among critically ill patients. We found that three different causal effects explained the disparity: (i) minority patients were admitted younger on average, decreasing the mortality rate; (ii) minority patients had worse chronic health and were more likely to be admitted for non-elective and medical reasons, increasing the mortality risk; (iii) there was a protective direct effect for minority group patients admitted for medical reasons, indicating a decrease in mortality compared to the majority group, when all other variables were kept equal. The last, protective direct effect may be explained through a causal mechanism in which minority patients are, due to worse access to primary health care, more likely to end up in ICU for less severe and preventable conditions, which then spuriously reduces the mortality risk.
Finally, we emphasize that the novel framework used in this manuscript can be applied across a range of problems in health equity, as it provides a systematic way of causally explaining health disparities, while being compatible with the modern tools of large-scale data analysis and causal inference.

\newpage
% declarations
\backmatter

%\bmhead{Supplementary information}

%\section*{Acknowledgements}

\section*{Declarations}

\noindent \textbf{Indigenous Data Statement.}
Indigenous data is that which is ``generated, intentionally or not, by,
about, or for Aboriginal and Torres Strait Islander people''. It
also refers to ``Information, in any format or medium, collected, analysed, stored, and interpreted within the context of Indigenous individuals, collectives, populations, entities, lifeways, cultures, knowledge systems, lands, biodiversity, water and other resource'' \citep{IndigenousData2024}. 
Given the history of exploitation of Indigenous people in research and data collection practices in Australia, it is therefore critical that Indigenous knowledges and approaches are prioritized when seeking or using data for research
from Indigenous populations \citep{kukutai2016indigenous, rainie2019indigenous, walker2019state}.

The datasets in this study that were obtained from ANZICS and contain
information relating to Aboriginal and Torres Strait Islander people in
Australia. Data has been used considering the international FAIR
principles of; Findable, Accessible, Interoperable and Reusable, and
CARE principles of; Collective benefit, Authority to control, Responsibility
and Ethics \citep{alliance2019care}. These principles will be embedded to
ensure that Indigenous data governance and Indigenous knowledges are
prioritized, but also ensure that the results benefit Indigenous peoples.

\begin{itemize}
\item \textbf{Funding:} Author LAC is funded by the National Institute of Health through R01 EB017205, DS-I Africa U54 TW012043-01 and Bridge2AI OT2OD032701, and the National Science Foundation through ITEST \#2148451. Other authors were funded by their respective institutions.
\item \textbf{Conflict of interest/Competing interests:} Authors declare no competing interests or conflicts of interest. 
\item \textbf{Ethics approval and consent to participate:} The research in this paper was approved by the ethics committee of the Alfred Hospital, Melbourne (Project \#661/24). This study utilized retrospective, de-identified data, and therefore, informed consent was waived by the ethics committee in accordance with applicable regulations and guidelines. Our study was also performed in collaboration with the Indigenous Data Network (IDN). 
\item \textbf{Consent for publication:} All authors consent to the publication of the manuscript.
\item \textbf{Data availability:} The MIMIC-IV dataset is publicly available from \href{https://physionet.org/content/mimiciv/3.1/}{Physionet} \citep{goldberger2000physiobank}. The ANZICS APD is available through the Australian and New Zealand Intensive Care Society (\href{https://www.anzics.org/adult-patient-database-apd/}{ANZICS}) upon request. 
\item Materials availability: N/A.
\item \textbf{Code availability:} All the code used in the paper is available in the Github repository \url{https://github.com/dplecko/RaceMortalityICU}. The README file provides instructions for installation and reproducing the results, and each script contains comments explaining the code logic.
\item \textbf{Author contributions:} DP, EB, LAC, and RB developed the study design. DP carried out the statistical analyses and data processing. DP wrote the manuscript. PS, LAC, and DPilcher helped revise the manuscript and interpret the results. AC, AF, SA, and DD provided discussions on the interpretation of the results and their implications for minority health.
\end{itemize}

\bibliographystyle{bst-files/sn-basic}
\bibliography{refs}
\newpage
\begin{appendices}
\section{Patient Information \& Filtering}\label{appendix:patient-info}
\renewcommand{\floatpagefraction}{1} 
\renewcommand{\textfraction}{0}
\renewcommand{\topfraction}{1}
\begin{samepage}
    \begin{figure}[h]
\begin{subfigure}[b]{0.425\textwidth}
        \centering
\resizebox{\linewidth}{!}{
    \begin{tikzpicture}[
    node distance=2.5cm and 2cm,
    box/.style={rectangle, draw, minimum width=4cm, minimum height=1.5cm, align=center},
    arrow/.style={-Stealth, thick},
]

% Nodes
\node[box] (step1) {n = 1,160,903 \\ (202 sites)};
\node[box, below=of step1] (step2) {n = 1,035,890 \\ (181 sites)};
\coordinate (midpoint) at ($(step1.south)!0.5!(step2.north)$);
\node[box, below=of step2] (step3) {ANZICS APD\\ Cohort};
\node[box, right=of midpoint, xshift=1cm] (filter1) {%
    \begin{tabular}{c}
    age $<$ 18 years (n = 10,848) \\[4pt]
    sex missing (n = 262) \\[4pt]
    elective status missing \\ (n = 4,860) \\[4pt]
    repeated admission \\ (n = 50,255) \\[4pt]
    Indigenous status missing \\
    (n = 58,788)
    \end{tabular}
};

% Arrows
\draw[arrow] (midpoint) -- (filter1.west);
\draw[arrow] (step1.south) -- (step2.north);
\draw[arrow] (step2.south) -- (step3.north);

\node[below=0.5cm of step3] {};
\end{tikzpicture}
}
        \caption{ANZICS APD filtering.}
        \label{fig:aics-filtering}
    \end{subfigure}
    \hfill
    \begin{subfigure}[b]{0.425\textwidth}
        \centering
        \resizebox{\linewidth}{!}{
\begin{tikzpicture}[
    node distance=2.5cm and 2cm,
    box/.style={rectangle, draw, minimum width=4cm, minimum height=1.5cm, align=center},
    arrow/.style={-Stealth, thick},
]

% Nodes
\node[box] (step1) {n = 73,181};
\node[box, below=of step1] (step2) {n = 38,844};
\coordinate (midpoint) at ($(step1.south)!0.5!(step2.north)$);
\node[box, below=of step2] (step3) {MIMIC-IV Cohort};
\node[box, right=of midpoint, xshift=1cm] (filter1) {%
    \begin{tabular}{c}
    repeated admission \\ (n = 22,261) \\[4pt]
    race missing or not \\ Caucasian / 
    African-American \\ (n = 12,076)
    \end{tabular}
};

% Arrows
\draw[arrow] (step1.south) -- (step2.north);
\draw[arrow] (step2.south) -- (step3.north);
\draw[arrow] (midpoint) -- (filter1.west);

\node[below=0.5cm of step3] {};
\end{tikzpicture}
}
        \caption{MIMIC-IV filtering.}
        \label{fig:miiv-filtering}
    \end{subfigure}
    \caption{Study flowchart of patient filtering steps for (a) ANZICS APD and (b) MIMIC-IV.}
    \label{fig:study-flowchart}
\end{figure}
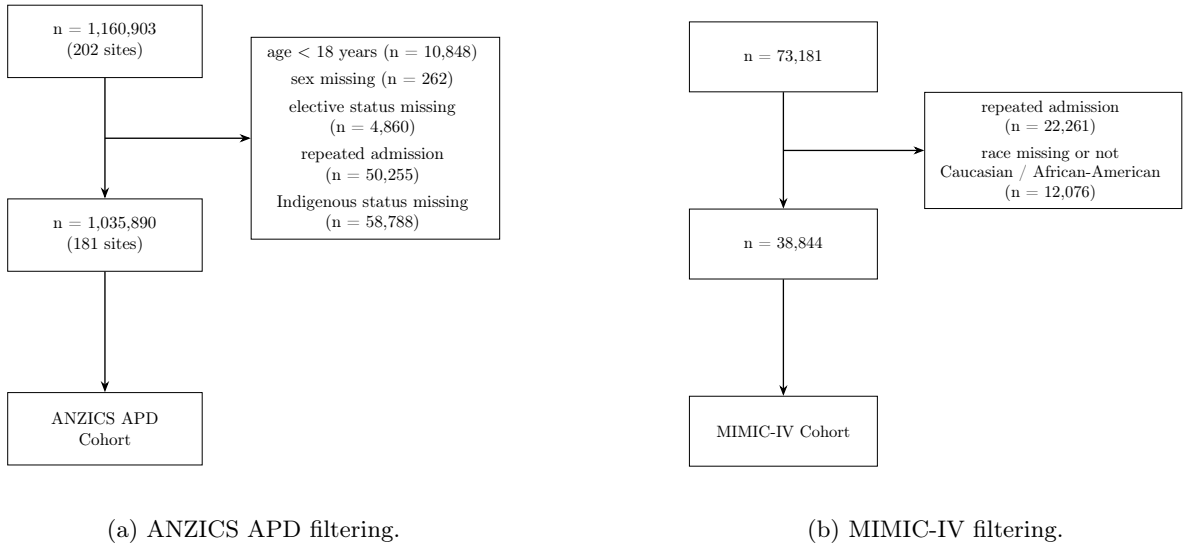

\vspace{-1em}
\begin{table}[h]
\begin{tabular}{lcccc}
\toprule
Variable & Reported & ANZICS APD (Majority) & ANZICS APD (Indigenous) & p-value\\
\midrule
Cohort size & n & 997154 & 38736 & \\
Age (years) & Median (IQR) & 66.58 (52.88-76.08) & 50.6 (37.09-61.6) & $ < 0.001$\\
Admission type &  &  &  & $ < 0.001$\\[2pt]
\quad - Medical & \% & 45 & 69 & \\
\quad - Surgical & \% & 55 & 31 & \\
\addlinespace
Mortality & n (\%) & 74820 (7.5\%) & 3064 (7.9\%) & 0.003\\
ICU LOS (days) & Median (IQR) & 1.76 (0.92-3.25) & 2.02 (1.01-3.96) & $ < 0.001$\\
Hospital LOS (days) & Median (IQR) & 7.71 (4.05-14.06) & 7.16 (3.48-14.2) & $ < 0.001$\\
% Sex (Female) & \% & 44 & 49 & \\
Sex (Male) & \% & 56 & 51 & $ < 0.001$\\
Ventilated & n (\%) & 306332 (30.7\%) & 15164 (39.1\%) &  $< 0.001$\\
APACHE-III Score & Mean, Med. (IQR) & 50.7, 47 (35-62) & 51.6, 47 (33-66) & 0.063\\
APACHE-III Risk of Death & Mean, Med. (IQR) & 0.12, 0.05 (0.02-0.14) & 0.14, 0.06 (0.02-0.16) & $< 0.001$ \\
\bottomrule
\end{tabular}
\caption{Comparison of patient characteristics on ANZICS APD.}
\label{tab:anzics-pts-tbl}
\end{table}

\begin{table}[h]
\centering
\begin{tabular}{lcccc}
\toprule
Variable & Reported & MIMIC-IV (White) & MIMIC-IV (African-American) & p-value\\
\midrule
Cohort size & n & 34204 & 4640 & \\
Age (years) & Median (IQR) & 67 (55-78) & 60 (47-72) & $ < 0.001$\\
Admission type &  &  &  & $ < 0.001$\\
\quad - Medical & \% & 60 & 75 & \\
\quad - Surgical & \% & 40 & 25 & \\
Mortality & n (\%) & 2627 (7.7\%) & 318 (6.9\%) & $ < 0.001$\\
ICU LOS (days) & Median (IQR) & 1.85 (1.08-3.37) & 1.8 (1-3.26) & $ < 0.001$\\
Hospital LOS (days) & Median (IQR) & 6.5 (3.91-10.9) & 6.41 (3.68-11.54) & 0.460\\
% Sex (Female) & \% & 44 & 54 & \\\addlinespace
Sex (Male) & \% & 56 & 46 & $ < 0.001$\\
Ventilated & n (\%) & 13692 (40.0\%) & 1480 (31.9\%) & $ < 0.001$\\
SOFA &  &  &  & $ < 0.001$\\
\quad- Respiratory & Median (IQR) & 2 (1-3) & 2 (1-3) & \\
\quad- Coagulation & Median (IQR) & 0 (0-1) & 0 (0-1) & \\
%\addlinespace
\quad- Hepatic & Median (IQR) & 0 (0-1) & 0 (0-1) & \\
\quad- Cardio & Median (IQR) & 1 (1-1) & 1 (0-1) & \\
\quad- CNS & Median (IQR) & 0 (0-1) & 0 (0-1) & \\
\quad- Renal & Median (IQR) & 0 (0-1) & 0 (0-1) & \\
\quad- Total & Median (IQR) & 4 (2-6) & 4 (2-6) & \\
\bottomrule
\end{tabular}
\caption{Comparison of patient characteristics on MIMIC-IV.}
\label{tab:mimic-pts-tbl}
\end{table}
\end{samepage}
\newpage
\section{Interaction Testing} \label{appendix:interaction-testing}

\begin{table}[h!]
\centering
\renewcommand{\arraystretch}{1.5}
\scalebox{1.3}{
\begin{tabular}{|c|c|c|}
\hline
\textbf{Interaction Test} & \textbf{ANZICS APD (AU)} & \textbf{MIMIC-IV (US)} \\ 
\hline
TE $\otimes$ SE           & $< 0.01^*$   & 0.11        \\\hline
DE $\otimes$ IE            & $0.02^*$   & $0.02^*$        \\\hline
DE $\otimes$ SE            & $0.04^*$         & 0.71        \\\hline
IE $\otimes$ SE            & $0.03^*$   & $< 0.01^*$  \\\hline
DE $\otimes$ IE $\otimes$ SE  & 0.59       & 0.11        \\\hline
% TE $\otimes$ SE (log-risk)          & 0.12         & $0.04^*$  \\\hline
% DE $\otimes$ IE (log-risk)           & 0.65         & 0.59        \\\hline
% DE $\otimes$ SE (log-risk)           & $0.04^*$   & 0.09        \\\hline
% IE $\otimes$ SE (log-risk)           & 0.78         & 0.69        \\\hline
% DE $\otimes$ IE $\otimes$ SE (log-risk) & 0.51       & 0.59        \\ 
% \hline
\end{tabular}
}
\captionsetup{width=\textwidth}
\vspace{0.1in}
\caption{Interaction Testing for ANZICS APD and MIMIC-IV datasets.}
\label{tab:interaction-testing}
\end{table}

\newpage
\section{Investigating Overlap} \label{appendix:overlap}
An important assumption required for correct causal effects estimation is known as overlap:
\begin{align} \label{eq:overlap}
    \delta < P(X = 1 \mid Z = z, W = w) < 1 - \delta,
\end{align}
for some $\delta > 0$. In this appendix, we investigate the overlap assumption in the datasets we analyzed. We abbreviate the quantity $P(X = 1 \mid Z = z, W = w)$ by $e_{x_1}(z,w)$, and similarly $P(X = 0 \mid Z = z, W = w)$ by $e_{x_0}(z, w)$. The minimum of the two propensity weights $e_{\min}(z,w)$ is defined as 
\begin{align}
    e_{\min}(z,w) := \min \{ e_{x_0}(z,w), e_{x_1}(z,w) \}.
\end{align}
The notation $e_x(Z,W)$ denotes a random variable depending on random values $Z, W$, as opposed to a fixed value $e_x(z,w)$ for $Z =z, W=w$.
To investigate the validity of the overlap assumption in Eq.~\ref{eq:overlap}, we first estimate the propensity weights $\hat e_x(Z_i,W_i)$ for each of our data samples $(Z_i, W_i)$. Then, we perform a sensitivity analysis using quantile-based trimming of the propensity weights \citep{crump2009dealing}. Let $Q(q; e_{\min}(Z,W))$ denote the lower $q$ quantile of the distribution of the propensity weights $e_{\min}(Z,W)$. For each dataset, and each quantile $q \in \{1\%, \dots, 5\%\}$, we select all the samples with values above the lower $q$ quantile, and construct the dataset
\begin{align}
   \mathcal{D}(q) = \{(X_i, Z_i, W_i, Y_i) \;\;\forall i:  \hat e_{\min}(Z_i,W_i) > Q(q; \hat e_{\min}(Z,W))\}.
\end{align}
We then perform the decomposition of the TV measure as in Eq.~\ref{eq:tv-decomp} for the subset of the data $\mathcal{D}(q)$ with the $q$-quantile of samples with extreme propensity weights removed. For the dataset $\mathcal{D}(q)$, for which the overlap criterion in Eq.~\ref{eq:overlap} should hold with the value of $\delta = Q(q; \hat e_{\min}(Z,W))$, we compute the 95\% O-value using the difference-in-tails method (DiT) \citep{lei2021distribution}, labeled $\hat{\mathcal{O}}(q)$, which provides an upper bound on the overlap value $\delta$ for the dataset $\mathcal{D}(q)$, in a distribution-free way. To test for overlap violations, we use the hypothesis test that compares if $\hat{\mathcal{O}}(q)$ is smaller than the nominal overlap bound $Q(q; \hat e_{\min}(Z,W))$ of the trimmed dataset $\mathcal{D}(q)$. If so, the test detects overlap violations and signals that a stronger trimming procedure is necessary.

\begin{figure}[h!]
    \centering
    \includegraphics[width=0.825\linewidth]{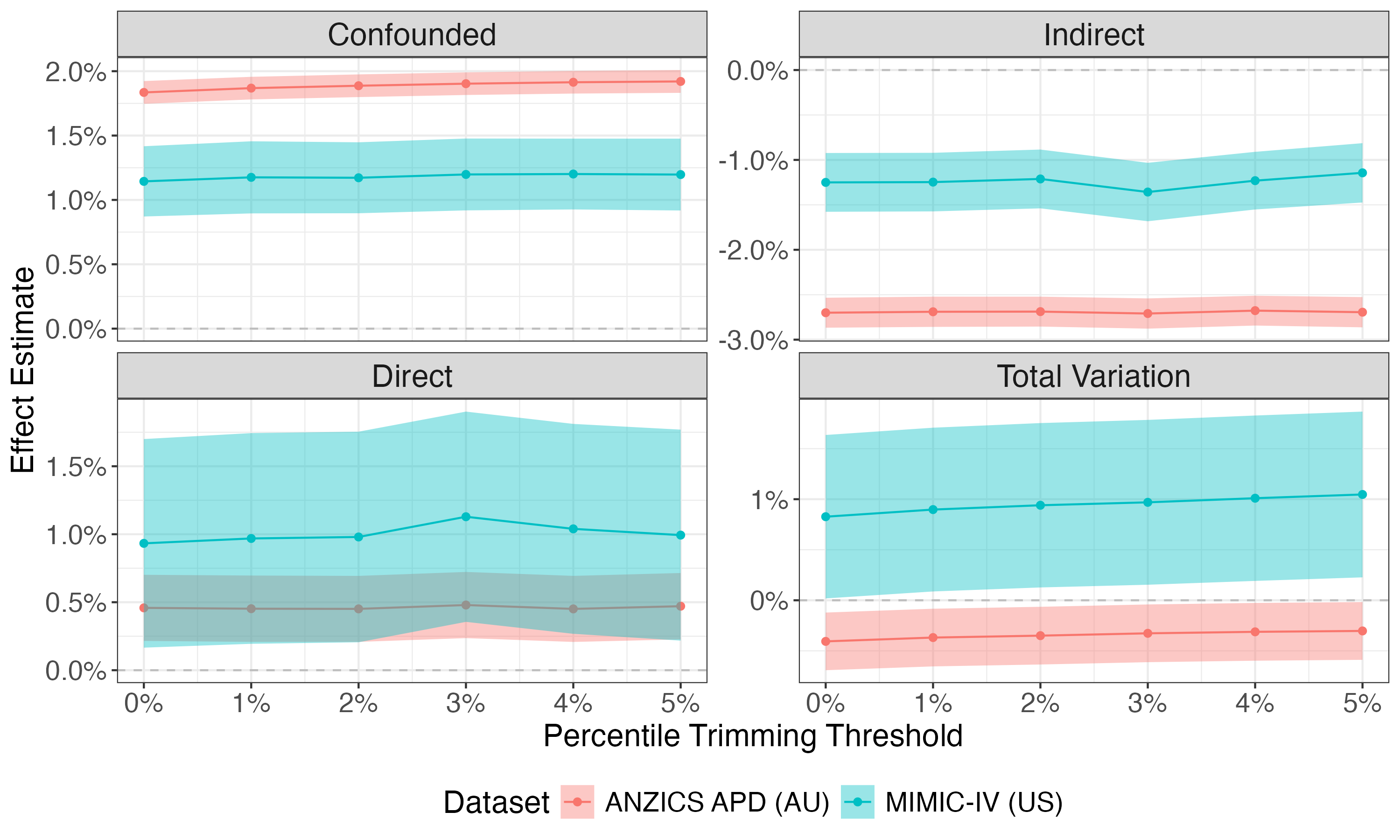}
    \caption{Sensitivity analysis of the impact of quantile-based trimming of propensity weights on TV decompositions on ANZICS APD and MIMIC-IV.}
    \label{fig:overlap-analysis}
\end{figure}
Fig.~\ref{fig:overlap-analysis} provides the results of the sensitivity analysis, and demonstrates that the causal decomposition is not significantly affected by the trimming of propensity weights over different thresholds $q$.
Furthermore, for both datasets and at each threshold $q$, we found that the O-values were greater than the nominal overlap condition after trimming, namely $\hat{\mathcal{O}}(q) > Q(q; \hat e_{\min}(Z,W))$, meaning that no significant overlap violations were detected (trimming was strong enough). Putting everything together, we conclude that there is no evidence that the TV decompositions are affected by overlap violations.

\newpage
\section{Indirect Effect Heterogeneity} \label{appendix:ie-hetero}
\begin{figure}[h]
    \centering
    \begin{subfigure}[b]{0.775\textwidth}
        \centering
        \includegraphics[width=0.8\textwidth]{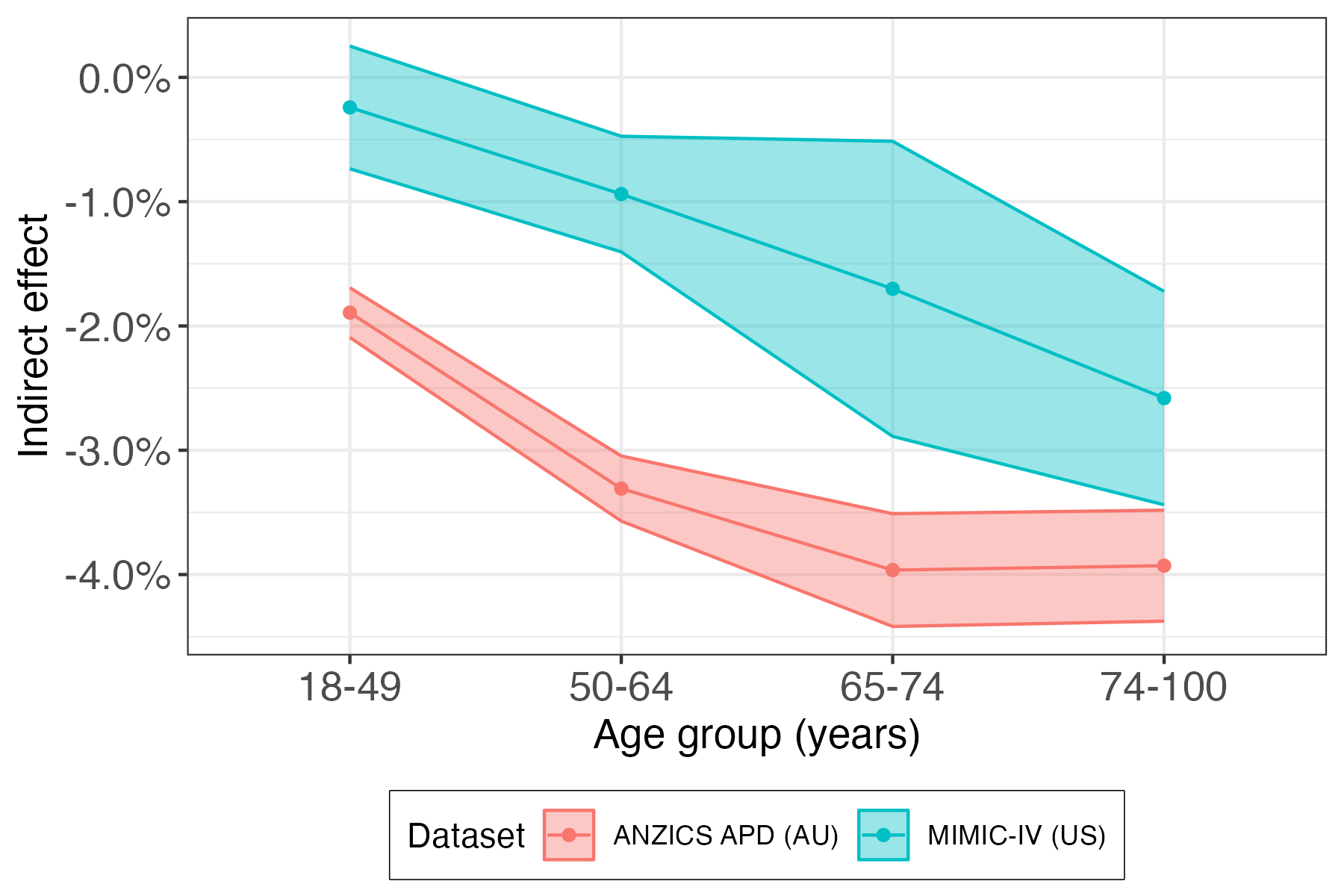}
        \caption{Indirect effect heterogeneity across age groups on ANZICS APD and MIMIC-IV.}
        \label{fig:ie-hetero}
    \end{subfigure}
    \hfill
    \begin{subfigure}[b]{0.48\textwidth}
        \centering
        \includegraphics[width=\textwidth]{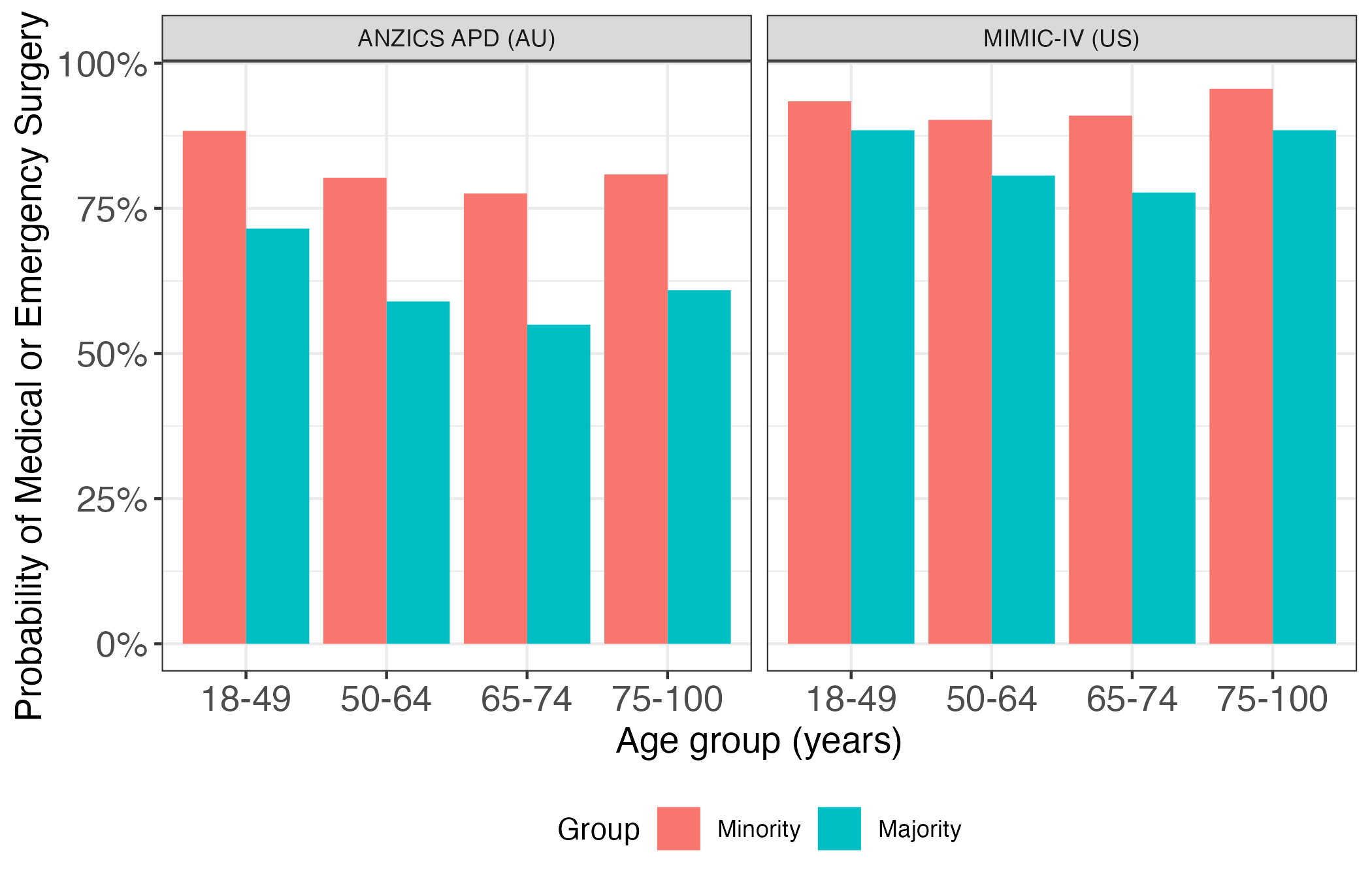}
        \caption{Effect of minority status on admission type.}
        \label{fig:minority-to-admission}
    \end{subfigure}
    \hfill
    \begin{subfigure}[b]{0.48\textwidth}
        \centering
        \includegraphics[width=\textwidth]{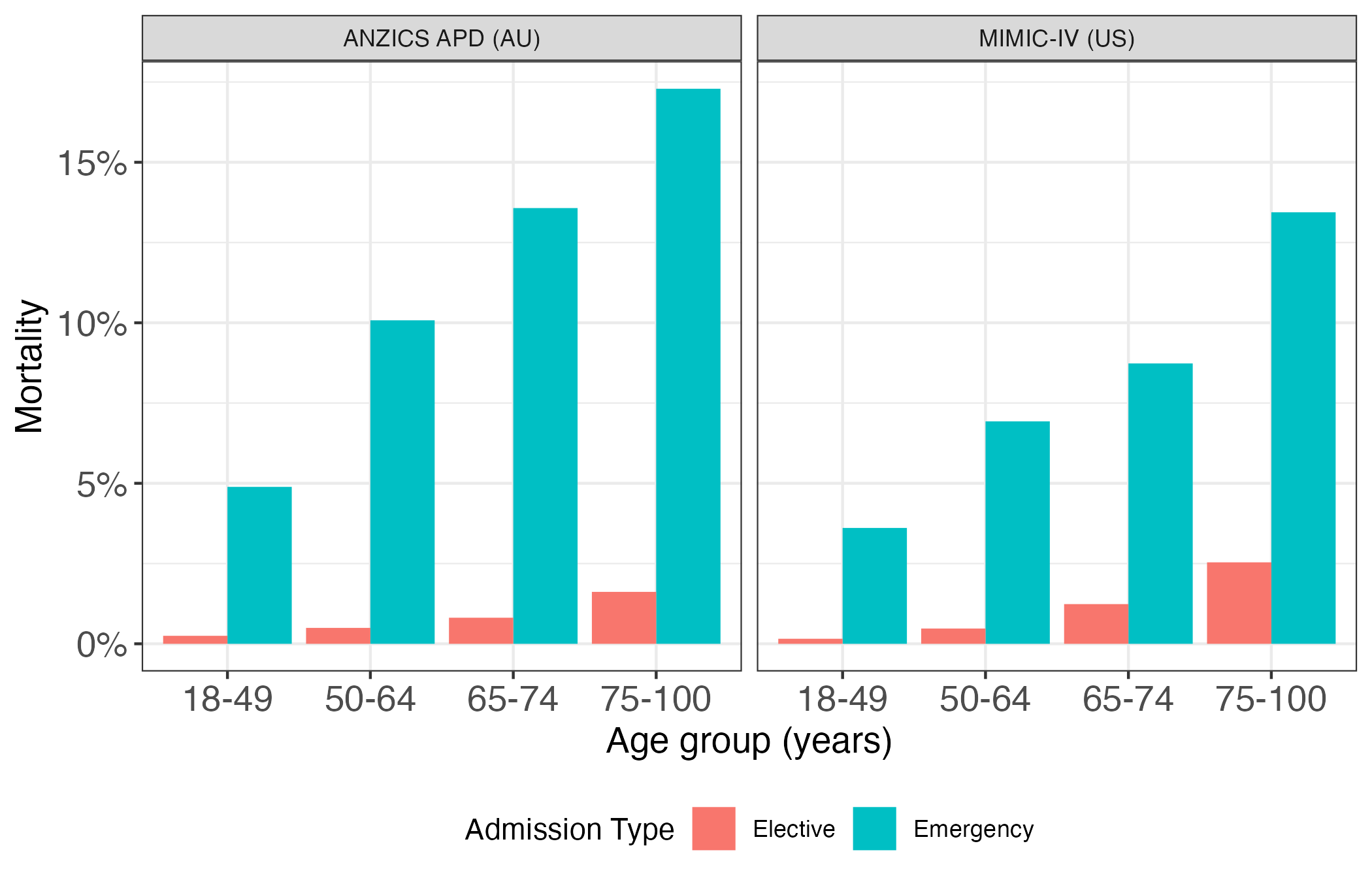}
        \caption{Effect of admission type on mortality.}
        \label{fig:admission-to-mortality}
    \end{subfigure}
    \caption{Understanding the heterogeneity of indirect effects on ANZICS APD and MIMIC-IV data.}
\end{figure}

The interaction testing analysis performed in Sec.~\ref{sec:interaction-testing} demonstrated that important interactions exist between indirect and spurious pathways. In this appendix we focus on the heterogeneity of indirect effects according to age, as a consequence of significant interactions of spurious and indirect paths. 
For the four age groups 18-54, 55-64, 65-74, and 75-100, labeled $C_1$ to $C_4$, and corresponding approximately to the quartiles of age in the data, we compute the indirect effects
\begin{align}
    -\text{IE}_{x_1, x_0}(y \mid \text{age} \in C_i, x_0) = \ex[Y_{x_1, W_{x_1}} - Y_{x_1, W_{x_0}} \mid \text{age} \in C_i, X = x_0].
\end{align}
The effects are estimated for both the Australian and US data, and visualized in Fig.~\ref{fig:ie-hetero}. In both datasets, the strength of the indirect increases with age, and the risk of death for majority group patients is reduced along the indirect path.
After establishing these effects, we explain mechanistically what drives this heterogeneity. In Fig.~\ref{fig:minority-to-admission}, we plot how the proportion of medical and emergency surgery admissions (these admissions are jointly referred to as urgent) changes across age groups $C_i$ and minority status. In Fig.~\ref{fig:admission-to-mortality}, we plot how the mortality rate for urgent and elective admissions changes across age groups $C_i$. Fig.~\ref{fig:minority-to-admission} illustrates that in every age group $C_i$, minority patients are more likely to be admitted for an urgent condition. Fig.~\ref{fig:admission-to-mortality} illustrates that while the risk of death increases with age for both urgent and elective admissions, the difference in risk of urgent vs. elective admissions becomes larger with age, making the indirect effect more pronounced. These observations explain why the indirect effect, along which minority patients are more likely to die, is more pronounced with older age.
\newpage
\section{Missing Value Sensitivity Analysis} \label{appendix:miss-sens}
\begin{figure}[h]
    \centering
    \includegraphics[width=0.75\linewidth]{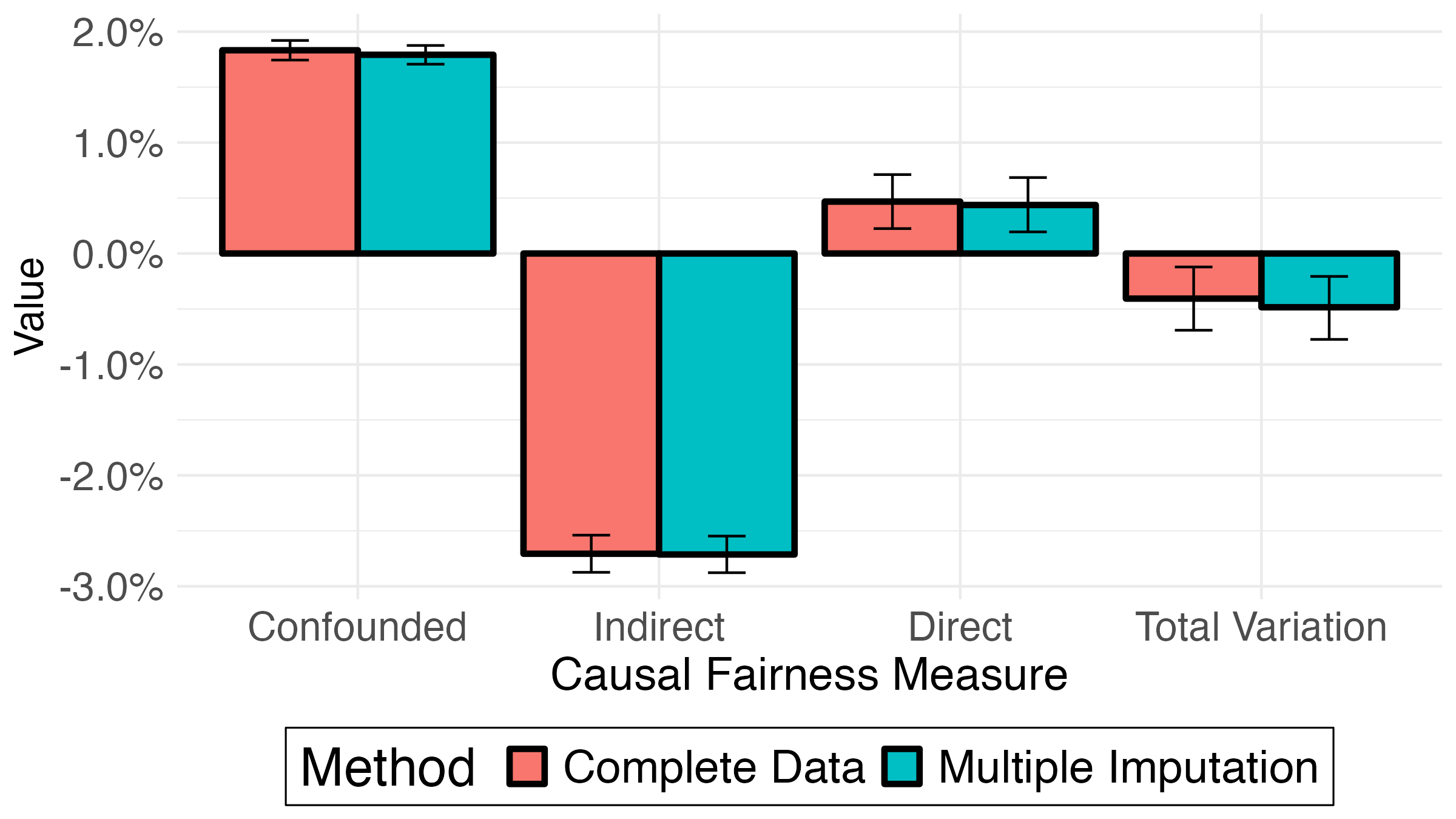}
    \caption{Sensitivity analysis of how the imputation of missing values of $X$ affects the TV decomposition on the ANZICS APD data.}
    \label{fig:miss-sens}
\end{figure}
\noindent In this appendix, we investigate the effect of missing values on the data analysis. In the ANZICS APD, of the overall cohort under investigation, $n= 58788$ patients (5.4\%) had no reported Indigenous status. To investigate the effect of this data missingness on the decompositions of the TV measure, we proceed as follows. 
Let $M$ denote the missingness indicator. We assume that the data is missing at random (MAR), meaning that the missingness pattern depends on the other observed variables $Z, W, Y$, and not on the value of $X$ itself, i.e.,
\begin{align}
    M \ci X \mid Z, W, Y.
\end{align}
The MAR condition can also be written as
\begin{align}
    P(M = 1 \mid X = x_1, Z=z, W=w, Y=y) = P(M = 1 \mid X = x_0, Z=z, W=w, Y=y).
\end{align}
For a fixed value of $Z=z, W=w, Y=y$, the missingness of $X$ does not depend on the actual value of $X$. A standard consequence of the MAR assumption is the fact that
\begin{align}
    P(X = x \mid M = 1, Z=z, W=w, Y=y) = P(X = x \mid M = 0, Z=z, W=w, Y=y).
\end{align}
In words, the distribution of $X$ given $Z, W, Y$ is the same within the subsets of data where $X$ is unobserved ($M=1$) and where $X$ is observed ($M=0$). Therefore, we perform the following sensitivity analysis. We first regress $X$ on $Z, W, Y$ using xgboost \citep{chen2016xgboost} on the complete subset of data with $X$ recorded. Then, for the subset where $X$ is missing, we impute the missing values of $X$ based on the observed values $Z=z,W=w, Y=y$. We repeat the imputation process ten times, obtaining imputed datasets $\mathcal{D}_1, \dots, \mathcal{D}_{10}$. We repeat the TV decomposition on each dataset $\mathcal{D}_i$, and compute the effect estimates and their confidence interval. These results (based on imputed data) are then compared to the data analysis focusing on complete data only, to understand how the effect estimates change, and how much uncertainty is added once missingness is taken into account. The results of this experiment are shown in Fig.~\ref{fig:miss-sens}, and indicate that the data missingness has a negligible impact on the results.

\end{appendices}

\end{document}